\pdfoutput=1

\documentclass{article}

\usepackage{microtype}
\usepackage{graphicx}
\graphicspath{{media/}}
\usepackage{subfigure}
\usepackage{booktabs}
\usepackage{hyperref}
\usepackage{xcolor}

\usepackage[accepted]{mlsys2025}
\makeatletter
\renewcommand{\mlsys@appearing}{Preprint. Under review.}
\makeatother
\hypersetup{pdfsubject={}}

\makeatletter
\g@addto@macro\normalsize{%
  \setlength\abovedisplayskip{6pt plus 2pt minus 3pt}%
  \setlength\belowdisplayskip{6pt plus 2pt minus 3pt}%
}
\makeatother

\makeatletter
\def\section{\@startsection{section}{1}{\z@}{-0.105in}{0.02in}{\large\bf\scshape\raggedright}}
\def\subsection{\@startsection{subsection}{2}{\z@}{-0.088in}{0.01in}{\normalsize\bf\raggedright}}
\def\subsubsection{\@startsection{subsubsection}{3}{\z@}{-0.07in}{0.01in}{\normalsize\itshape\raggedright}}
\def\paragraph{\@startsection{paragraph}{4}{\z@}{1.3ex plus 0.5ex minus .2ex}{-1em}{\normalsize\bf}}
\makeatother

\usepackage{amsmath,amsfonts,bm}

\def\eqref#1{equation~\ref{#1}}

\def\1{\bm{1}}

\DeclareMathAlphabet{\mathsfit}{\encodingdefault}{\sfdefault}{m}{sl}
\SetMathAlphabet{\mathsfit}{bold}{\encodingdefault}{\sfdefault}{bx}{n}

\usepackage[utf8]{inputenc} %
\usepackage{lipsum} %

\usepackage[T1]{fontenc}

\usepackage{etoolbox}
\newbool{includeappendix}
\setbool{includeappendix}{true}

\ifdefined\isoverfull
\else
\fi

\usepackage{xcolor} %

\definecolor{my-full-blue}{HTML}{1F77B4}

\definecolor{my-full-orange}{HTML}{FF7F0E}

\definecolor{my-full-green}{HTML}{2CA02C}

\definecolor{my-full-red}{HTML}{d62728}

\definecolor{my-full-purple}{HTML}{9467bd}

\colorlet{my-blue}{my-full-blue!30}
\colorlet{my-orange}{my-full-orange!30}
\colorlet{my-green}{my-full-green!30}
\colorlet{my-red}{my-full-red!30}
\colorlet{my-purple}{my-full-purple!30}

\usepackage{url}

\usepackage{xcolor}
\usepackage{xspace}
\usepackage{enumitem}
\usepackage{amsmath,amssymb}
\usepackage{mathtools}
\usepackage{array}
\usepackage{multirow}
\usepackage{makecell}
\usepackage{float}
\usepackage{siunitx}
\usepackage{colortbl}

\usepackage{amsthm}

\theoremstyle{plain}
\newtheorem{theorem}{Theorem}[section]
\newtheorem{proposition}[theorem]{Proposition}
\newtheorem{lemma}[theorem]{Lemma}
\newtheorem{corollary}[theorem]{Corollary}
\theoremstyle{definition}

\theoremstyle{remark}

\newcommand{\ours}{\textit{our method}\xspace}
\newcommand{\Ours}{\textit{Our method}\xspace}%
\newcommand{\ourslabel}{\textit{Ours}\xspace}%

\makeatletter\newcommand{\alglabel}[1]{\protected@edef\@currentlabel{\arabic{ALC@line}}\label{#1}}\makeatother

\usepackage[capitalize]{cleveref}

\crefformat{section}{\S#2#1#3}

\crefrangeformat{section}{\S#3#1#4\crefrangeconjunction\S#5#2#6}

\crefmultiformat{section}{\S#2#1#3}{\crefpairconjunction\S#2#1#3}{\crefmiddleconjunction\S#2#1#3}{\creflastconjunction\S#2#1#3}

\newcommand{\crefrangeconjunction}{--}

\crefname{listing}{Lst.}{listings}
\crefname{line}{Lin.}{Lin.}
\crefname{appendix}{App.}{App.}
\crefname{subappendix}{App.}{App.}
\crefname{subsubappendix}{App.}{App.}

\newcommand{\app}[1]{%
	\ifbool{includeappendix}{\cref{#1}}{the appendix}%
}
\newcommand{\App}[1]{%
	\ifbool{includeappendix}{\cref{#1}}{The appendix}%
}

\newcommand{\speedupForward}{1.8--6.4$\times$}

\newcommand{\speedupBackward}{2.0--21$\times$}
\newcommand{\speedupLosses}{T5's backward pass at 128 and 256 tokens (0.75--0.82$\times$)}

\newcommand{\decompositionForward}{1.1--2.2$\times$}

\newcommand{\decompositionBackward}{1.7--228$\times$}

\newcommand{\memoryForward}{2.9--128$\times$}

\newcommand{\memoryBackward}{1.4--54$\times$}

\newcommand{\ablationAccumulationLatencyBackward}{2.9--4.4$\times$}
\newcommand{\ablationAccumulationMemoryBackward}{1.8--2.6$\times$}
\newcommand{\ablationProjectionLatencyForward}{2.0--2.9$\times$}

\newcommand{\ablationProjectionMemoryForward}{0.19--0.43$\times$}

\newcommand{\ablationRetentionLatencyBackward}{1.1$\times$}
\newcommand{\ablationRetentionMemoryBackward}{0.42--0.50$\times$}
\newcommand{\ablationTargetedLatency}{1.3$\times$}
\newcommand{\ablationTargetedMemory}{1.5$\times$}
\newcommand{\trafficAccumulation}{1.3--3.1$\times$}
\newcommand{\trafficAccumulationTXL}{3.1$\times$}
\newcommand{\trafficRowGroupsTXL}{1.4$\times$}
\newcommand{\trafficRowGroupsDeBERTa}{1.0$\times$}
\newcommand{\trafficConstantRanges}{0.95--1.0$\times$}
\newcommand{\trafficExpansionDeBERTa}{1.2--1.8$\times$}
\newcommand{\trafficProjection}{0.044--0.19$\times$}
\newcommand{\parakeetSegmentsSpeedupSDPA}{1.2$\times$}

\newcommand{\parakeetTalksSpeedupSDPA}{1.8$\times$}
\newcommand{\parakeetTalksMemorySDPA}{2.4$\times$}

\newcommand{\canarySpeedup}{1.09--1.36$\times$}

\newcommand{\canaryLossGap}{0.4\%}

\newcommand{\debertaAccuracyGap}{0.3}
\newcommand{\debertaFOneGapShort}{+1.3}
\newcommand{\debertaFOneGapRest}{0.3}

\mlsystitlerunning{Fast and Exact Relative-Position Attention via Structure-Aware Kernels}

\begin{document}

\twocolumn[
  \mlsystitle{Fast and Exact Relative-Position Attention\texorpdfstring{\\}{ }via Structure-Aware Kernels}

  \begin{mlsysauthorlist}
    \mlsysauthor{Delyan Boychev}{insait}
    \mlsysauthor{Ivo Petrov}{insait}
    \mlsysauthor{Dimitar I. Dimitrov}{insait}
  \end{mlsysauthorlist}

  \mlsysaffiliation{insait}{INSAIT, Sofia University ``St. Kliment Ohridski''}

  \mlsyscorrespondingauthor{Delyan Boychev}{delyan.boychev@insait.ai}

  \mlsyskeywords{efficient attention, GPU kernels, relation-aware attention, Triton, training systems}

  \vskip 0.3in

  \begin{abstract}
  Relative-position attention mechanisms offer compelling alternatives to RoPE across various transformer architectures, delivering superior performance in domains such as audio. These mechanisms typically modify attention scores by adding content--position or bias--position terms. Unlike RoPE, which encodes position via per-token rotations directly on the queries and keys, relative-position terms generally require pairwise offset-dependent computation or lookups, complicating their integration into fused attention kernels and introducing severe memory, bandwidth, and compute overheads as sequences scale. To address this bottleneck, we propose an exact attention algorithm that incorporates additive relative-position terms into FlashAttention-style tiled computation. \Ours (1) uses offset-aware tiling to avoid redundant indexing and memory accesses, (2) computes only the necessary key/query--relation products and reuses them across token pairs in both forward and backward passes, and (3) uses offset-based gradient sharding and aggregation, eliminating atomic contention. We provide a unified, efficient Triton implementation across diverse relative-position attention mechanisms and evaluate it against naive and specialized implementations. For 8K-token sequences, \ours speeds up the forward (backward) pass by \speedupForward{} (\speedupBackward{}) and reduces peak memory up to \memoryForward{} (\memoryBackward{}) over the strongest baseline for each mechanism, enabling long-context end-to-end transcription that is \parakeetTalksSpeedupSDPA{} faster and uses \parakeetTalksMemorySDPA{} less memory. \Ours thus provides a drop-in, hardware-efficient solution for scaling relative-position attention transformers.
\end{abstract}

]

\printAffiliationsAndNotice{}

\section{Introduction}
\label{sec:introduction}

The Transformer architecture~\citep{attentionisallyouneed} has revolutionized modern machine learning, from large language models (LLMs) in natural language processing to foundation models in speech, vision, and biology. At its core is the scaled dot-product attention (SDPA) mechanism, which allows input tokens to directly attend to each other and aggregate mutual information. However, scaling these models to larger inputs introduces two fundamental challenges, namely injecting scalable positional information into the network, and keeping the computation and memory footprint of SDPA tractable despite its quadratic complexity.

\begin{figure*}[t]
  \centering
  \includegraphics[width=\textwidth]{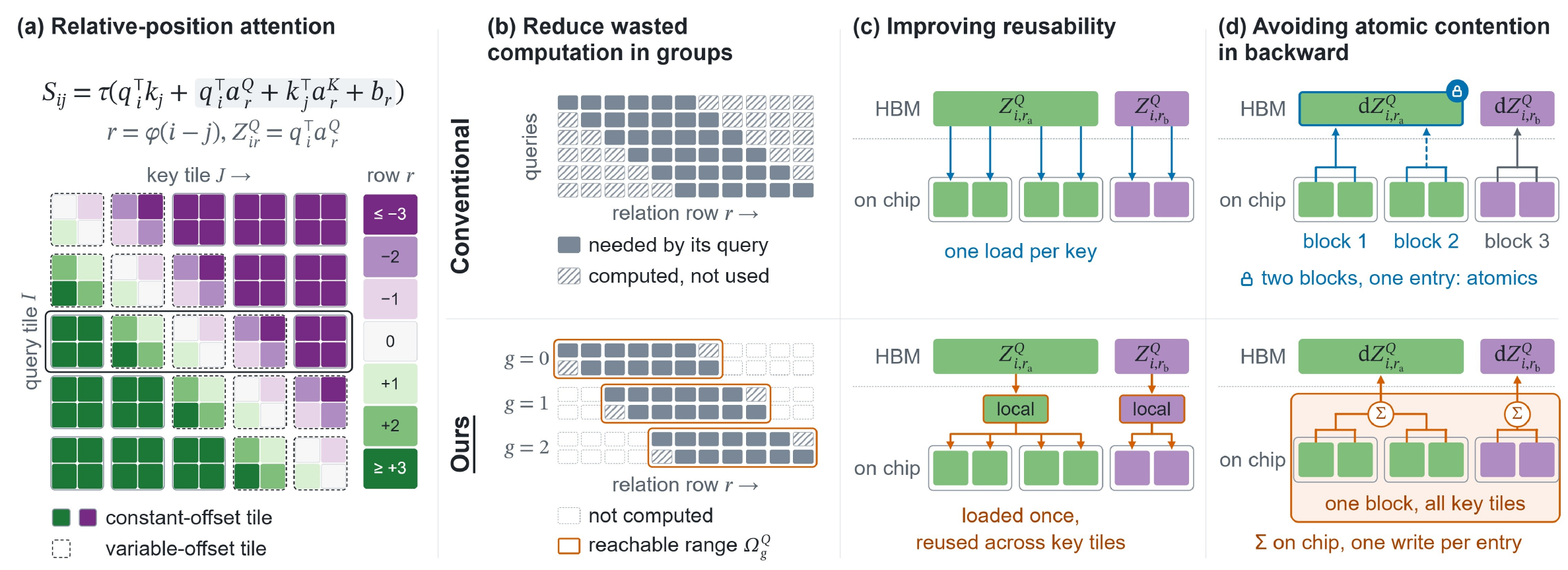}%
  \caption{Overview of \ours. \textbf{(a)} Each score adds the relation terms of its row $r=\phi(i,j)$, which stays constant on tiles far from the diagonal. \textbf{(b)} \Ours first computes each query group $g$ only on its reachable rows $\Omega^Q_g$. \textbf{(c)} Each shared entry $Z^Q_{i,r}$ is loaded once and reused across key tiles. \textbf{(d)} One thread block per query tile sums gradients on chip, avoiding atomic contention.}
  \label{fig:overview}
\end{figure*}

For standard content-based attention, IO-aware kernels such as FlashAttention~\citep{flashattention,flashattention2} resolved the quadratic memory bottleneck by streaming blocks of tokens through fast on-chip SRAM, avoiding the materialization of the full $N \times N$ score matrix in GPU high-bandwidth memory (HBM). Paired with parameter-free, token-wise rotary position embeddings (RoPE)~\citep{rope}, which can be seamlessly integrated into fused kernels, this pipeline now serves as the backbone of most autoregressive LLMs.

\paragraph{Relative-position attention}
Parameter-free positional encodings, such as RoPE and ALiBi~\citep{alibi}, are not without drawbacks. Despite their computational efficiency, they often fail to use context beyond the sequence lengths seen during training~\citep{kazemnejad2023impact,fope}, where learned relative biases such as T5's generalize better~\citep{kazemnejad2023impact,fire}. Consequently, learnable relative-position encodings remain an enticing alternative for Transformer models. This class of mechanisms additively augments attention scores with learnable terms that depend on the relative distance $i-j$ between query and key tokens, either through content-to-position projections or scalar biases. This allows modeling explicit, shift-invariant relationships across sequences, and has remained enticing for speech models~\citep{seamless,canary,phi4}, text encoding~\citep{debertav3,gliner2}, proteomics~\citep{prottrans}, and genomics~\citep{borzoi}.

\paragraph{Efficiency limitations of relative-position attention}

Despite their modeling advantages, relative-position attention mechanisms are largely incompatible with standard fused attention kernels. Training or running inference with these models typically requires materializing the full $N \times N$ additive position matrix in HBM, reintroducing the exact memory-bandwidth bottleneck that FlashAttention was designed to eliminate. Programmable attention frameworks such as FlexAttention~\citep{flexattention} offer a partial solution by fusing user-defined per-score callbacks and enabling limited precomputation; however, they still suffer from redundant computation, repeated memory lookups, and atomic contention during backpropagation. Lacking efficient kernels, practitioners often resort to replacing relative-position encodings entirely and retraining models from scratch, as seen in Flashzoi~\citep{flashzoi} and FlashSwin~\citep{flashswin}, or switch to RoPE-based alternatives such as ModernBERT~\citep{modernbert}.

\paragraph{Our work}

To address these limitations, we introduce an exact algorithm for relative-position mechanisms. Rather than treating pairwise offsets as opaque per-element lookups, \ours exploits the block-level structure of relative distances across the attention matrix (\cref{fig:overview}a). Specifically, \ours projects queries and keys only onto the offset ranges reachable within each token block (\cref{fig:overview}b), structures tile traversal around constant- and variable offset regions to maximize data reuse and avoid redundant indexing (\cref{fig:overview}c), and shards backward gradient accumulation by offset to eliminate atomic contention entirely (\cref{fig:overview}d).

Empirically, \ours substantially reduces both runtime and memory consumption across forward and backward passes in production speech and text models.  Our kernels significantly outperform generalist and specialized implementations, as well as state-of-the-art frameworks such as FlexAttention, FlashAttention-4~\citep{flashattention4}, and FlashInfer~\citep{flashinfer}, even when those frameworks are augmented with our preprojected relation decomposition. In particular, for $8\text{K}$-token sequences, \ours beats the strongest baseline for each mechanism by \speedupForward{} in the forward pass and \speedupBackward{} in the backward pass, while reducing peak memory usage by \memoryForward{} and \memoryBackward{}, respectively. Applied end-to-end to long-context speech transcription, it delivers a \parakeetTalksSpeedupSDPA{} speedup and \parakeetTalksMemorySDPA{} lower memory consumption on Parakeet-TDT-0.6B-v3~\citep{canary}, enabling significantly more efficient deployment and training.

\paragraph{Key contributions} Our main contributions include:

\begin{itemize}\setlength{\itemsep}{0pt}
  \item A unified formulation of additive relative-position attention spanning four mechanisms, exposing shared structure amenable to kernel-level optimization.
  \item An exact, drop-in tiled algorithm for fast and memory-efficient relative-position attention.
  \item High-performance implementations of \ours, validated against a high-precision reference, and comprehensive evaluations across diverse model architectures, sequence lengths, and NVIDIA GPU architectures.
\end{itemize}

\section{Related Work}
\label{sec:related-work}
In this section, we describe relevant prior work on optimizing standard attention and programmable backends for non-standard attention mechanisms. We then describe the uses of relative-position attention and their limitations due to the lack of general optimization algorithms.

\subsection{Optimizing Standard Attention}
Modern Transformer architectures rely on scaled dot-product attention (SDPA)~\citep{attentionisallyouneed}. However, naive implementations materialize the full attention score and probability matrices in high-bandwidth memory (HBM), incurring storage and I/O costs that grow quadratically with sequence length. To mitigate this bottleneck, \citet{milakov2018online} introduced an online softmax formulation that processes inputs in column-wise chunks, enabling exact attention computation in sub-quadratic memory~\citep{selfattentiondoesnotneed}. FlashAttention~\citep{flashattention} combines this with IO-aware tiling in on-chip memory, recomputing attention tiles during the backward pass instead of storing large intermediate matrices. FlashAttention-2~\citep{flashattention2} restructures the schedule to improve parallelism and reduce overhead by parallelizing the forward pass over query blocks and the backward pass over key blocks. FlashAttention-3 and -4~\citep{flashattention3,flashattention4} further specialize this pipeline for newer GPUs. 

Existing kernels target content-based attention and cannot fuse learnable relative-position terms or their gradients. \Ours bridges this gap by extending tiled execution to relative-position attention by restricting projections to reachable offsets, reusing them on constant-offset tiles, and using a two-pass backward schedule~\citep{triton2024fusedattention} for atomic-free gradient accumulation.

\subsection{Programmable and Specialized Backends}
To support attention variants without hand-writing a kernel per model, recent frameworks expose programmable interfaces that preserve FlashAttention's optimizations. FlexAttention~\citep{flexattention} compiles user-defined mask predicates and elementwise score modifications into fused tiled kernels, automatically deriving their backward pass. FlashInfer~\citep{flashinfer}, AttentionEngine~\citep{attentionengine}, and Flashlight~\citep{flashlight} provide similar customization through templates or compilers, while others target variant families such as sparse masks~\citep{flashmask} or low-rank biases~\citep{flashbias}. 

Because these frameworks express score modifications as pointwise callbacks on query--key pairs, they can only support relative-position terms via elementwise lookups into precomputed tables. Because this abstraction hides the tile structure, it cannot exploit spatial locality across constant-offset pairs, leading to redundant memory accesses during the forward pass and contention-heavy atomic additions~\citep{flexattention} during gradient accumulation.

\subsection{Relative-Position Attention}

\paragraph{Applications of relative-position attention}
While modern autoregressive language models largely rely on fixed positional schemes like RoPE~\citep{rope} and ALiBi~\citep{alibi} that integrate neatly into standard SDPA kernels, many domains benefit from expressive, learnable relative-position encodings that inject pairwise token distances directly into the attention logits. Introduced by \citet{shawrelative}, this approach projects queries onto learned embeddings of clipped relative offsets. Transformer-XL~\citep{transformerxl} expands this notion by incorporating unclipped offsets with learned query biases, DeBERTa~\citep{deberta,debertav3} by modelling both content-to-position and position-to-content interactions, and T5~\citep{t5} by mapping offsets to learned scalar bucket biases. 
Today, these mechanisms remain foundational for long-context modeling across speech, text, and biology. They support the backbone of conformer-based~\citep{conformer} speech and multimodal models such as Parakeet, Canary, SeamlessM4T~v2, and Phi-4-multimodal~\citep{canary,seamless,phi4}, DeBERTa-based text and extraction pipelines~\citep{gliner2}, and biological sequence models in proteomics and genomics~\citep{prottrans,enformer,borzoi}. \Ours unifies the common computational patterns across all these architectures, providing a single hardware-efficient algorithm.

\paragraph{Optimizing relative-position attention}
Despite their widespread adoption, these architectures remain difficult to scale due to the overhead of computing and applying relative-position terms. Early implementations reduced the cost of positional projections by evaluating query- and key-relation dot products once per unique offset, and then mapping the resulting table via explicit index lookups~\citep{deberta,irpe} or row-shifting reshape operations~\citep{musictransformer,transformerxl}. However, both strategies still materialize a dense quadratic bias matrix in HBM.  
Subsequent systems have attempted to fuse this pipeline, though they remain either incomplete or model-specific.  For instance, TensorRT's DeBERTa plugin~\citep{nvidia2022tensorrtdeberta} and NeMo's Transformer-XL path~\citep{kuchaiev2019nemo} fuse the offset-gather step yet still materialize the full score matrix in HBM. On the other hand, FlashDeBERTa~\citep{knowledgator2025flashdeberta} and GLiNER4j~\citep{gravitee2026gliner4j} integrate positional terms directly into tiled FlashAttention-style kernels (the latter for inference only), but are strictly tailored to DeBERTa. Lacking general kernel support, practitioners often modify and retrain the model instead: Flashzoi~\citep{flashzoi} replaces relative-position encodings with RoPE, with analogous replacements appearing for speech recognition~\citep{ropeasr} and proposed for vision Transformers~\citep{flashswin,lee2026rib}. Others turn to kernelized attention to incorporate relative positions in sub-quadratic time~\citep{luo2021stable,liutkus2021spe}, but sacrificing exactness, which has unclear performance impact. \Ours accelerates both training and inference across the full family of relative-position architectures without compromising model fidelity or numerical exactness.

\section{Background}
\label{sec:background}
In this section, we introduce the necessary background for understanding our work. We first recall the definition of standard SDPA and how FlashAttention-1~\citep{flashattention} and -2~\citep{flashattention2} optimize it. We then define a unified notion of relative--position attention, and explain the drawbacks of existing implementations. We include a full table of the notations used in our paper in \cref{app:notation} for convenience.
 
\subsection{Standard Efficient Attention}
We consider a self-attention block for a single sequence of length $N$ and a single attention head of dimension $d$, since batches and heads are computed independently. It takes query, key, and value matrices $Q, K, V \in \mathbb{R}^{N \times d}$ with row vectors $q_i^\top, k_j^\top, v_j^\top$ ($i, j \in \{0, \dots, N-1\}$) and computes the output $O \in \mathbb{R}^{N \times d}$ with rows $o_i^\top$:
\begin{equation}
  \label{eq:standard-sdpa}
  S = \tau Q K^\top + M, \qquad P = \operatorname{softmax}_{\text{row}}(S), \qquad O = P V.
\end{equation}
Here $\tau$ is a normalization scalar, typically $d^{-\frac{1}{2}}$, $M \in \{0, -\infty\}^{N\times N}$ is the additive attention mask, $S, P \in \mathbb{R}^{N \times N}$ are the complete softmax logits and row-normalized probabilities, respectively, and $\operatorname{softmax}_{\text{row}}$ operates row-wise: $P_{ij} = \exp(S_{ij}) / \sum_{t=0}^{N-1} \exp(S_{it})$.

\paragraph{Backward pass} For any tensor or scalar $X$, we denote the gradient of the training loss with respect to $X$ as $dX$. Given the incoming output gradient $dO \in \mathbb{R}^{N \times d}$, the backward pass computes the input gradients $dQ, dK, dV \in \mathbb{R}^{N \times d}$. Differentiating $O = P V$ yields $dV = P^\top dO$ and $dP = dO V^\top$, which propagate to $dS = P \odot (dP - D \mathbf{1}^\top)$, where $D \in \mathbb{R}^N$ has entries $D_i = \langle dO_i, o_i \rangle$. Finally, differentiating $S = \tau Q K^\top + M$ gives $dQ = \tau dS K$ and $dK = \tau dS^\top Q$.

\paragraph{Tiled execution}
Materializing the intermediate matrices $S$ and $P$ in HBM requires $O(N^2)$ workspace, which causes slow reads and writes, even though neither is needed once $O$ is produced. FlashAttention-1 and -2 eliminate this bottleneck by partitioning sequences into query tiles $\mathcal{I}$ and key tiles $\mathcal{J}$---sets of consecutive positions processed together in on-chip SRAM. It parallelizes the forward pass over query tiles: each thread block loads one query tile $Q_{\mathcal{I}}$ and iterates over key and value tiles $(K_{\mathcal{J}}, V_{\mathcal{J}})$. For each query, it keeps a running maximum $m_i$, denominator sum $\ell_i$, and unnormalized output $u_i \in \mathbb{R}^d$. At each step $\mathcal{J}$, it computes the score tile $S_{\mathcal{I}\mathcal{J}} = \tau Q_{\mathcal{I}} K_{\mathcal{J}}^\top + M_{\mathcal{I}\mathcal{J}}$ on-chip and updates the state via online softmax~\citep{milakov2018online}:
\begin{equation}
  \label{eq:online-softmax}
  \begin{aligned}
    m_i' &= \max\!\big(m_i, \max_{j \in \mathcal{J}} S_{ij}\big), \quad
    \ell_i' = e^{m_i - m_i'} \ell_i + \sum_{j \in \mathcal{J}} e^{S_{ij} - m_i'}, \\
    u_i' &= e^{m_i - m_i'} u_i + \sum_{j \in \mathcal{J}} e^{S_{ij} - m_i'} v_j.
  \end{aligned}
\end{equation}
After all key tiles $\mathcal{J}$ are processed, the worker writes the normalized output $o_i = u_i / \ell_i$ and saves one scalar row log-normalizer per query $L_i = m_i + \log \ell_i$. 

In the backward pass, FlashAttention avoids the $O(N^2)$ memory overhead of storing $P$ by trading on-chip recomputation for reduced HBM traffic. It takes the saved $(Q, K, V, O, L)$ and incoming $dO$, precomputes $D_i = \langle dO_i, o_i \rangle$, and accumulates input gradients across tiles via $dV_{\mathcal{J}} \mathrel{+}= P_{\mathcal{I}\mathcal{J}}^\top dO_{\mathcal{I}}$, $dK_{\mathcal{J}} \mathrel{+}= \tau dS_{\mathcal{I}\mathcal{J}}^\top Q_{\mathcal{I}}$, and $dQ_{\mathcal{I}} \mathrel{+}= \tau dS_{\mathcal{I}\mathcal{J}} K_{\mathcal{J}}$ in $O(N)$ workspace. Because $P$ was not stored, each tile $(\mathcal{I}, \mathcal{J})$ recomputes $S_{\mathcal{I}\mathcal{J}}$ on-chip and uses $L_{\mathcal{I}}$ to reconstruct $P_{\mathcal{I}\mathcal{J}}$ and $dS_{\mathcal{I}\mathcal{J}}$ on the fly:
\begin{equation}
  \label{eq:prob-reconstruction}
  \begin{aligned}
    P_{\mathcal{I}\mathcal{J}} &= \exp\!\big(S_{\mathcal{I}\mathcal{J}} - L_{\mathcal{I}} \mathbf{1}^\top\big), \\
    dS_{\mathcal{I}\mathcal{J}} &= P_{\mathcal{I}\mathcal{J}} \odot \big(dO_{\mathcal{I}} V_{\mathcal{J}}^\top - D_{\mathcal{I}} \mathbf{1}^\top\big).
  \end{aligned}
\end{equation}

\subsection{Relative-Position Attention}
\label{sec:background-relative}
Relative-position attention extends standard SDPA by incorporating the relative positions of queries and keys directly into their attention scores. Specifically, it adds terms that measure the interaction of the query or key content vectors with a representation of their relative distance and/or assign a scalar bias based on that distance. We express these terms using up to three relation tables with $R$ rows ($r \in \{0, \dots, R-1\}$): query- and key-relation vector tables $A^Q, A^K \in \mathbb{R}^{R \times d}$ with rows $(a^Q_r)^\top, (a^K_r)^\top$, and a scalar-bias table $b \in \mathbb{R}^R$. These tables may be learned directly as parameters or computed by applying learned linear projections to position embeddings. A discrete map $\phi(i, j) \in \{0, \dots, R-1\}$ then assigns each query--key pair $(i, j)$ to its corresponding relation-row index $r$.

For common offset-based mechanisms, $\phi(i, j)$ depends only on the offset $\delta = i - j$, with rows ordered by increasing $\delta$. Standard mappings fall into three categories:
\begin{itemize}\setlength{\itemsep}{0pt}
	\item \textbf{Identity}: each integer offset $-N+1\leq\delta\leq N-1$ has its own row $r=\phi(i, j) = \delta + N-1$, so $R = 2N - 1$.
	\item \textbf{Clipped}: like identity, but offsets are clamped to $[-\Delta, \Delta]$ for a chosen maximum offset $\Delta$. In particular, $r=\phi(i, j) = \operatorname{clip}(\delta, -\Delta, \Delta) + \Delta$ and $R = 2\Delta + 1$.
	\item \textbf{Bucketed}: the most general case: any $\phi(i,j)$ that splits the offsets $\delta$ into $R$ consecutive, ordered buckets.
\end{itemize}

All such mechanisms share a single canonical score equation that replaces the logits in \cref{eq:standard-sdpa} with
\begin{equation}
	\label{eq:canonical-score}
	S_{ij} = \tau \left( q_i^\top k_j + q_i^\top a^Q_{\phi(i,j)} + k_j^\top a^K_{\phi(i,j)} + b_{\phi(i,j)} \right) + M_{ij},
\end{equation}
where some relation terms can be absent per mechanism. Four widely deployed mechanisms fit this unified view:
\begin{itemize}\setlength{\itemsep}{0pt}
	\item \textbf{Shaw}~\citep{shawrelative} uses $A^Q$, a \emph{clipped} map, and sets $\tau = d^{-\frac{1}{2}}$.
	\item \textbf{Transformer-XL (TXL)}~\citep{transformerxl} uses $A^Q$ and $b$, an \emph{identity} map, and sets $\tau = d^{-\frac{1}{2}}$.
	\item \textbf{DeBERTa}~\citep{deberta,debertav3} uses $A^Q$, $A^K$, a \emph{clipped or bucketed} map, and sets $\tau = (3d)^{-\frac{1}{2}}$.
	\item \textbf{T5}~\citep{t5} uses only $b$ under a logarithmic \emph{bucketed} map.
\end{itemize}

For notational convenience, we define the query- and key-relation terms for each token and relation row $r$, as follows:
\begin{equation}
	\label{eq:token-relation-projections}
	Z^Q_{ir} = q_i^\top a^Q_r \quad \text{and} \quad
	Z^K_{jr} = k_j^\top a^K_r,
\end{equation}
forming the matrices $Z^Q, Z^K \in \mathbb{R}^{N \times R}$. For each query–key pair $(i,j)$, the corresponding terms in \cref{eq:canonical-score} are then the entries at $r=\phi(i,j)$ in $Z^Q$ and $Z^K$.

\paragraph{Backward pass}
Extending standard attention, the forward pass additionally takes $(A^Q, A^K, b)$ and $\phi$, while the backward pass must also compute the table gradients $(dA^Q, dA^K, db)$ alongside the additional positional contributions to $dQ$ and $dK$. Given the score gradients $dS$, differentiating through $r = \phi(i, j)$ gives:
\begin{equation}
	\label{eq:projection-gradients}
	dZ^Q_{ir} = \tau \!\!\sum_{j:\,\phi(i,j)=r}\!\! dS_{ij} \quad \text{and} \quad
	dZ^K_{jr} = \tau \!\!\sum_{i:\,\phi(i,j)=r}\!\! dS_{ij}.
\end{equation}
Differentiating the projections then updates the input gradients as $dQ \mathrel{+}= dZ^Q A^Q$ and $dK \mathrel{+}= dZ^K A^K$ compared to SDPA, and produces the relation-table gradients $dA^Q = (dZ^Q)^\top Q$ and $dA^K = (dZ^K)^\top K$ and $db = (dZ^Q)^\top \mathbf{1}$.

\subsection{Limitations of Existing Attention Backends}
\label{sec:background-limitations}
Existing implementations gather or shift copies of the relative-position terms into a dense square matrix before the softmax.
Specifically, eager implementations follow one of two strategies, both of which reintroduce quadratic HBM bottlenecks. Naive implementations instantiate $a^Q_{\phi(i,j)}$ for every $(i, j)$ pair, costing $O(N^2 d)$ work and $O(N^2 d)$ workspace. Preprojected implementations instead compute $Z^Q, Z^K \in \mathbb{R}^{N \times R}$ in $O(NRd)$ work and $O(NR)$ space, then gather or shift them into an $N \times N$ bias matrix. Even with tiling instead of the $N \times N$ bias, the precomputed tables require $O(NR)$ storage, which often include unreachable columns (wasting compute and memory), fetches the same entry multiple times (e.g., when offsets are clipped), and remains quadratic for identity maps.

In the backward pass, tiled execution faces an additional many-to-one reduction bottleneck in \cref{eq:projection-gradients}. Whenever multiple pairs $(i, j)$ map to the same row $r = \phi(i, j)$, up to $O(N)$ keys per query and $O(N^2)$ pairs overall, their score gradients $\tau dS_{ij}$ must be summed into the same $dZ^Q_{ir}$, $dZ^K_{jr}$, and $db_r$ entries, causing severe atomic contention when accumulated via per-pair updates. While programmable backends such as FlexAttention can avoid the dense $N \times N$ bias through fused per-score callbacks, a scalar callback only sees a single $(i, j)$ pair at a time. It cannot change how outer projections are allocated, reuse entries across a tile of pairs sharing the same row $r$, or coordinate gradient ownership in backward to avoid atomic additions. 
\Ours overcomes all of these bottlenecks with its relation-aware tiled approach. Derivation of all equations is given in \cref{app:derivations}.

\section{Method}
\label{sec:relaccel}
To address the limitations of \cref{sec:background-limitations}, \ours brings relative-position terms into FlashAttention-style tiled attention. It computes only the relation projections the tiles read (\cref{sec:relaccel-projections}) and accumulates their gradients by offset (\cref{sec:relaccel-gradients}), which together form its forward and backward passes (\cref{sec:relaccel-alg}).

\subsection{Reusing Relation Projections}
\label{sec:relaccel-projections}
For any offset-based relation map $\phi$, each positional term for a query--key pair $(i,j)$ depends only on the relation row $r=\phi(i,j)$ and either query $i$ or key $j$. Consequently, all $N^2$ pairwise positional contributions are determined by the $N\times R$ token--relation projections $Z^Q$ and $Z^K$ (\cref{eq:token-relation-projections}) and the scalar bias $b \in \mathbb{R}^R$. Rather than expanding these tables into a dense $N\times N$ bias matrix in HBM, \ours evaluates them on chip within each attention tile:
\begin{equation}
  \label{eq:projected-relation-scores}
  S_{ij} = \tau\bigl(q_i^\top k_j + Z^Q_{i,\phi(i,j)} + Z^K_{j,\phi(i,j)}
  + b_{\phi(i,j)}\bigr) + M_{ij}.
\end{equation}
Each attention logit thus requires at most three scalar lookups from $Z^Q$, $Z^K$, and $b$ rather than a $d$-dimensional dot product. We next eliminate the two remaining bottlenecks identified in \cref{sec:background-limitations} -- computing unreachable projection columns and performing redundant elementwise lookups.

\paragraph{Reachable Projection Ranges}
Although an identity map defines $2N-1$ relation rows, any single query $i$ only attends to $N$ keys. It thus accesses $N$ contiguous offsets $\delta \in [i-N+1, i]$, leaving the remaining $N-1$ rows of $A^Q$ unused. To omit these unreachable projections while preserving batched matrix multiplications, \ours partitions $Q$ into contiguous groups $Q_g$ of $G$ queries. Because consecutive queries access overlapping offsets, an entire group $Q_g$ reaches a contiguous range $\Omega^Q_g=\{gG,\ldots,gG+N+G-2\}$ of only $N+G-1$ rows, reducing overhead to only $G-1$ additional rows. \Ours projects each $Q_g$ onto $A^Q_{\Omega^Q_g}$ (and analogically for $K_g$ and $A^K_{\Omega^K_g}$) as a single dense matrix product. For $N \gg G$, this restriction nearly halves both projection work and HBM storage while remaining exact. The scalar bias $b$ requires no projection and is left unrestricted.

\paragraph{Constant-Range Reuse}
For clipped and bucketed maps with a small saturation limit $\Delta \ll N$, individual token groups already span nearly all $R$ relation rows, offering limited savings from range restriction. Instead, these maps saturate outside the diagonal band $|\delta| < \Delta$, mapping all offsets $\delta \ge \Delta$ to row $r = R-1$ and all $\delta \le -\Delta$ to row $r = 0$. To handle bucketed maps in the same way, \ours expands their $R$ buckets into $2\Delta+1$ rows by duplicating each bucket across the offsets it covers, replacing index lookups with clipping. In any tile where $|\delta| \ge \Delta$, all pairs map to the same row $r$. \Ours therefore loads a single entry per query ($Z^Q_{ir}$), one per key ($Z^K_{jr}$), and one bias $b_r$, broadcasting them across the tile. Only pairs in the diagonal band $|\delta| < \Delta$ require individual lookups, and their fraction $(2\Delta-1)/N$ approaches zero as $N$ grows. Under causal masking ($j \le i$), the situation is similar, but queries access only offsets $\delta \ge 0$ and saturate only for $\delta \ge \Delta$.

\subsection{Atomic-free Relation Gradient Accumulation}
\label{sec:relaccel-gradients}
In the backward pass, accumulating score gradients $\tau\,dS_{ij}$ directly into $dZ^Q$ and $dZ^K$  causes atomic contention whenever multiple pairs map to the same relation row $r$. However, while a fixed query $i$ may map many keys $j$ to the same row $r$, each key corresponds to a unique offset $\delta = i - j$. Building on the standard two-pass backward schedule~\citep{triton2024fusedattention}, \ours assigns each query tile to a single thread block and accumulates its gradients by offset $\delta$ into $d\tilde Z^Q$ rather than by row $r$. A symmetric pass over key tiles accumulates $d\tilde Z^K$. Because each entry has a single writer, this schedule eliminates all $O(N^2)$ atomic additions and fixes the summation order, ensuring determinism.

\paragraph{Local Boundary Sums}
Similar to the forward pass, writing a separate gradient entry for every offset $\delta$ is unnecessary in the saturated regions. Since all $\delta \ge \Delta$ map to row $R-1$ and all $\delta \le -\Delta$ map to row $0$, each thread block maintains two on-chip running sums per token, one for each boundary row, and writes them to HBM only once. Combined with the reachable ranges $\Omega^Q_g$ and $\Omega^K_g$, each token stores at most $\min(N+G-1,\, 2\Delta+1)$ gradient entries (identity maps have $\Delta=N-1$) instead of one per offset, $2N-1$. As in \cref{sec:relaccel-projections}, this nearly halves the gradient buffer for identity maps ($N \gg G$), while for clipped and bucketed maps the fraction $(2\Delta+1)/(2N-1)$ approaches zero as $N$ grows.

\paragraph{Relation-Row Reduction}
The stored offset gradients directly yield $dZ^Q$ and $dZ^K$ (\cref{eq:projection-gradients}). For identity and clipped maps, each stored column already corresponds to its own row $r$ ($d\tilde Z^Q = dZ^Q$). For bucketed maps, the duplicated columns of each bucket are summed back into that bucket's row in a fixed, deterministic order. The remaining gradients then reduce to matrix products, as in \cref{sec:background-relative}.
For $\Delta \ll N$, these reductions require $O(N\Delta d)$ work rather than $O(N^2 d)$ for per-pair accumulation. \Ours uses this atomic-free path whenever each row $r$ spans contiguous offsets, and falls back to atomic additions otherwise.

\subsection{Complete Algorithm}
\label{sec:relaccel-alg}
We now compose the end-to-end algorithm, designing both the forward and backward passes in a tiled manner. 

\subsubsection{Forward algorithm}
To avoid materializing dense attention matrices in HBM, \ours streams query and key tiles through on-chip SRAM, similarly carrying over FlashAttention's tiled schedule, which remains applicable as each score tile $S_{\mathcal{I}\mathcal{J}}$ depends only on the preprojected relation entries of its own queries $\mathcal{I}$ and keys $\mathcal{J}$. \Cref{alg:forward} extends standard tiled attention in two places: precomputing the group projections before the tiled loop (lines~\ref{line:copy}--\ref{line:project}) and adding the relative-position terms on chip within it (lines~\ref{line:corners}--\ref{line:add}).

\begin{algorithm}[t]
\caption{Forward pass of \ours for one head.}
\label{alg:forward}
\small
\begin{algorithmic}[1]
\REQUIRE $Q,K,V,A^Q,A^K,b$ in HBM, map $\phi$
\ENSURE $O,L,Z^Q,Z^K$ in HBM
\IF{\alglabel{line:copy} $\phi$ is bucketed}
  \STATE Copy bucket rows
\ENDIF
\STATE \alglabel{line:project} $Z^Q_g\gets Q_g(A^Q_{\Omega^Q_g})^\top$, $Z^K_g\gets K_g(A^K_{\Omega^K_g})^\top$ for all $g$
\FOR{\alglabel{line:fwd-loop} each query tile $\mathcal I$ \textbf{in parallel}}
  \STATE Load $Q_{\mathcal I}$ from HBM to on-chip SRAM
  \STATE On chip, initialize $m_i,\ell_i,u_i\gets-\infty,0,0$
  \FOR{each key tile $\mathcal J$}
    \STATE Load $K_{\mathcal J},V_{\mathcal J}$ from HBM to on-chip SRAM
    \STATE \alglabel{line:scores} On chip, compute $S_{\mathcal I\mathcal J}=\tau Q_{\mathcal I}K_{\mathcal J}^\top+M_{\mathcal I\mathcal J}$
    \STATE \alglabel{line:corners} $\delta_{\min}\gets\min\mathcal I-\max\mathcal J$, $\delta_{\max}\gets\max\mathcal I-\min\mathcal J$
    \IF{\alglabel{line:test} $\delta_{\min}\ge\Delta$ \textbf{or} $\delta_{\max}\le-\Delta$}
      \STATE \alglabel{line:saturated} $r\gets R-1$ or $0$ for the whole tile
    \ELSE
      \STATE \alglabel{line:band} $r\gets\phi(i,j)$ for each score
    \ENDIF
    \STATE \alglabel{line:add} On chip, $S_{ij}\mathrel{+}=\tau(Z^Q_{ir}+Z^K_{jr}+b_r)$ (pointwise)
    \STATE On chip, update $m_i,\ell_i,u_i$ by \cref{eq:online-softmax}
  \ENDFOR
  \STATE On chip, compute $o_i=u_i/\ell_i$ and $L_i=m_i+\log\ell_i$
  \STATE \alglabel{line:fwd-write} Write $O_{\mathcal I}$ and $L_{\mathcal I}$ to HBM
\ENDFOR
\end{algorithmic}
\end{algorithm}

Inside the inner loop, the edge indices of a score tile determine its extreme offsets $\delta_{\min}$ and $\delta_{\max}$, so two comparisons suffice to check whether the tile saturates (lines~\ref{line:corners}--\ref{line:test}). Saturated tiles load and broadcast one entry per query and key (line~\ref{line:saturated}), whereas tiles intersecting the diagonal band perform per-score lookups (line~\ref{line:band}). We align the group size with the maximum tile size, chosen to be $G=128$, ensuring that every tile reads its projections from a single group with only $1.6\%$ column overhead at $N=8\text{K}$ tokens. When a single group already reaches all $R$ rows, as for clipped and bucketed maps with $\Delta \ll N$, \ours uses one group for the whole sequence.
By default, the forward pass saves $Z^Q$ and $Z^K$ in HBM for the backward pass, though they can instead be recomputed during backpropagation to reduce activation memory, which would, however, double projection work.

\subsubsection{Complete Backward Algorithm}
The backward pass (\cref{alg:backward}) executes a query-parallel pass for $dQ$ and $d\tilde Z^Q$, followed by a key-parallel pass for $dK$, $dV$, and $d\tilde Z^K$. Both passes recompute $S_{\mathcal{I}\mathcal{J}}$ as in the forward pass (line~\ref{line:bwd-rebuild}), recover $P_{\mathcal{I}\mathcal{J}}$ and $dS_{\mathcal{I}\mathcal{J}}$ via \cref{eq:prob-reconstruction}, and accumulate $\tau\,dS_{ij}$ by offset with on-chip boundary sums for $|\delta| \ge \Delta$ (lines~\ref{line:bwd-qacc} and~\ref{line:bwd-kacc}). Once the tiled loops finish, lines~\ref{line:bwd-fold}--\ref{line:bwd-reduce} reduce duplicated bucket columns into $dZ^Q, dZ^K$ when applicable and compute the remaining gradients via matrix products (\cref{sec:background-relative}).

\begin{algorithm}[t]
\caption{Backward pass of \ours for one head.}
\label{alg:backward}
\small
\begin{algorithmic}[1]
\REQUIRE $Q,K,V,O,L,dO$ in HBM\\ $Z^Q,Z^K,A^Q,A^K,b$ in HBM, map $\phi$
\ENSURE $dQ,dK,dV,dA^Q,dA^K,db$ in HBM
\STATE \alglabel{line:bwd-delta} $D_i\gets\langle dO_i,o_i\rangle$ for all $i$
\FOR{\alglabel{line:bwd-qloop} each query tile $\mathcal I$ \textbf{in parallel}}
  \STATE Load $Q_{\mathcal I},dO_{\mathcal I},L_{\mathcal I},D_{\mathcal I}$ to on-chip SRAM
  \STATE On chip, initialize $dQ_{\mathcal I}$ and the entries of $\mathcal I$ to $0$
  \FOR{each key tile $\mathcal J$}
    \STATE Load $K_{\mathcal J},V_{\mathcal J}$ from HBM to on-chip SRAM
    \STATE \alglabel{line:bwd-rebuild} On chip, rebuild $S_{\mathcal I\mathcal J}$ as in Alg.~\ref{alg:forward}, lines~\ref{line:scores}--\ref{line:add}
    \STATE On chip, compute $P_{\mathcal I\mathcal J}$ and $dS_{\mathcal I\mathcal J}$ by \cref{eq:prob-reconstruction}
    \STATE On chip, $dQ_{\mathcal I}\mathrel{+}=\tau\,dS_{\mathcal I\mathcal J}K_{\mathcal J}$
    \STATE \alglabel{line:bwd-qacc} On chip, accumulate $\tau\,dS_{ij}$ by offset for each $i$
  \ENDFOR
  \STATE \alglabel{line:bwd-qwrite} Write $dQ_{\mathcal I}$ and the entries of $\mathcal I$ to HBM
\ENDFOR
\FOR{\alglabel{line:bwd-kloop} each key tile $\mathcal J$ \textbf{in parallel}}
  \STATE Load $K_{\mathcal J},V_{\mathcal J}$ from HBM to on-chip SRAM
  \STATE On chip, initialize $dK_{\mathcal J},dV_{\mathcal J}$, entries of $\mathcal J$ to $0$
  \FOR{each query tile $\mathcal I$}
    \STATE Load $Q_{\mathcal I},dO_{\mathcal I},L_{\mathcal I},D_{\mathcal I}$ to on-chip SRAM
    \STATE On chip, rebuild $S_{\mathcal I\mathcal J}$, $P_{\mathcal I\mathcal J}$, $dS_{\mathcal I\mathcal J}$ as above
    \STATE On chip, $dV_{\mathcal J}\mathrel{+}=P_{\mathcal I\mathcal J}^\top dO_{\mathcal I}$
    \STATE On chip, $dK_{\mathcal J}\mathrel{+}=\tau\,dS_{\mathcal I\mathcal J}^\top Q_{\mathcal I}$
    \STATE \alglabel{line:bwd-kacc} On chip, accumulate $\tau\,dS_{ij}$ by offset for each $j$
  \ENDFOR
  \STATE \alglabel{line:bwd-kwrite} Write $dK_{\mathcal J},dV_{\mathcal J}$ and the entries of $\mathcal J$ to HBM
\ENDFOR
\IF{\alglabel{line:bwd-fold} $\phi$ is bucketed}
  \STATE Sum bucket copies into $dZ^Q,dZ^K$
\ENDIF
\STATE \alglabel{line:bwd-qreduce} $dQ\mathrel{+}=dZ^QA^Q$, $dA^Q\gets(dZ^Q)^\top Q$, $db\gets(dZ^Q)^\top\mathbf 1$
\STATE \alglabel{line:bwd-reduce} $dK\mathrel{+}=dZ^KA^K$, $dA^K\gets(dZ^K)^\top K$
\end{algorithmic}
\end{algorithm}

\subsubsection{Implementation}
\label{sec:implementation}
We implement \ours in Triton~\citep{triton} as a drop-in replacement for relative-position attention, with adapters for Hugging Face and NeMo models that require no weight modifications. Each call dispatches a kernel specialized at compile time to the active relation terms, map type and $\Delta$, masking, sequence packing, dropout, precision, and tile sizes, avoiding runtime branching and any overhead from disabled terms. Tile sizes are selected by sequence-length bucket from lookup tables tuned offline per GPU architecture, falling back to a hardware resource heuristic when no table entry exists. Similarly, the saturation check (lines~\ref{line:corners}--\ref{line:test} of \cref{alg:forward}) splits each tile's inner loop into three contiguous ranges on the host rather than branching per tile. Sequence lengths remain runtime arguments so lengths within a bucket share compiled kernels, with an optional precompilation step to warm the compiler cache. Users can trade compute for memory by selecting whether the forward pass saves $Z^Q$ and $Z^K$ in HBM or recomputes them during backpropagation. Finally, all kernels are custom PyTorch operators compatible with \texttt{torch.compile}~\citep{ansel2024pytorch2}.

\section{Evaluation}
\label{sec:evaluation}

In this section, we present the empirical evaluation of \ours. We first describe our experimental setup, including the target mechanisms and hardware, alongside the baselines we compare against (\cref{sec:eval-setup}). We then evaluate \ours at two levels. At the attention-block level we benchmark forward and backward throughput and memory footprints (\cref{sec:eval-attention}). Second, we apply \ours in end-to-end tasks, including training and inference, while comparing downstream task quality across three production workloads (\cref{sec:eval-e2e}). Finally, we isolate and measure the impact of each algorithmic optimization of \ours in ablation studies (\cref{sec:eval-ablations}).

\subsection{Experimental Setup}
\label{sec:eval-setup}

\begin{table}[t]
\centering\footnotesize
\caption{The systems we compare against and their support for each relative-position mechanism. ``F'' means forward, ``B'' backward, ``\textdagger'' the backward pass fails our FP32 check from 4K tokens, ``--'' unsupported.}
\label{tab:systems}
\vspace{0pt}
\setlength{\tabcolsep}{1.5pt}
\resizebox{\columnwidth}{!}{%
\begin{tabular}{@{}lcccc@{}}
\toprule
 & Shaw & TXL & DeBERTa & T5 \\
\midrule
\multicolumn{5}{l}{\textbf{General (dense bias)}} \\[1pt]
PyTorch SDPA & \textsf{F+B} & \textsf{F+B} & -- & \textsf{F+B} \\
FlexAttention \citep{flexattention} & \textsf{F+B} & \textsf{F+B} & \textsf{F+B}\rlap{\smash{\textsuperscript{\textdagger}}} & \textsf{F+B} \\
FlashAttention-4 \citep[FA4;][]{flashattention4} & \textsf{F} & \textsf{F} & \textsf{F} & \textsf{F} \\
FlashInfer \citep{flashinfer} & \textsf{F} & \textsf{F} & \textsf{F} & \textsf{F} \\
\midrule
\multicolumn{5}{l}{\textbf{Specialized}} \\[1pt]
NeMo FlexAttention \citep{kuchaiev2019nemo} & -- & \textsf{F+B} & -- & -- \\
FlashDeBERTa \citep{knowledgator2025flashdeberta} & -- & -- & \textsf{F+B}\rlap{\smash{\textsuperscript{\textdagger}}} & -- \\
GLiNER4j \citep{gravitee2026gliner4j} & -- & -- & \textsf{F} & -- \\
\midrule
\multicolumn{5}{l}{\textbf{+ our decomposition}} \\[1pt]
FlexAttention & \textsf{F+B} & \textsf{F+B} & \textsf{F+B} & \textsf{F+B} \\
NeMo FlexAttention & -- & \textsf{F+B} & -- & -- \\
FA4, FlashInfer, Flashlight \citep{flashlight} & \textsf{F} & \textsf{F} & \textsf{F} & \textsf{F} \\
\midrule
PyTorch eager, compiled \citep{ansel2024pytorch2} & \textsf{F+B} & \textsf{F+B} & \textsf{F+B}\rlap{\smash{\textsuperscript{\textdagger}}} & \textsf{F+B} \\
\textbf{\ourslabel} & \textsf{F+B} & \textsf{F+B} & \textsf{F+B} & \textsf{F+B} \\
\bottomrule
\end{tabular}}
\end{table}

\paragraph{Mechanisms}
We evaluate the four mechanisms of \cref{sec:background-relative} with the maps their models use. Shaw clips offsets, as in the w2v-BERT encoders of SeamlessM4T \citep{seamless}, here at $\Delta=256$. TXL uses the identity map, as in the FastConformer encoders of Canary and Parakeet \citep{fastconformer,canary}. DeBERTa uses 256 logarithmic buckets up to distance 512, as in DeBERTa-v3 \citep{debertav3}, and T5 uses 32 bidirectional buckets up to distance 128, as in the original T5 \citep{t5}.

\begin{figure*}[t]
  \centering
  \includegraphics[width=\textwidth]{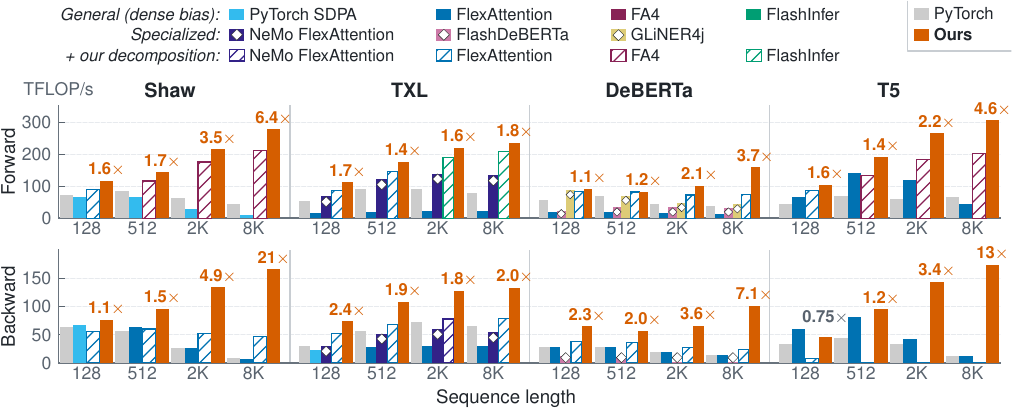}
  \caption{Attention throughput on an H200 (FP16). PyTorch, general, and decomposition bars show the fastest system of their kind; labels give the speedup of \ours over the fastest system without our decomposition.}
  \label{fig:kernels}
\end{figure*}

\paragraph{Baselines}
We compare to general and mechanism-specific implementations in \cref{tab:systems}, from their official repositories, Hugging Face Transformers \citep{wolf2020transformers}, or NeMo, with every dense bias built by the models' attention modules, as in practice. We call those without our decomposition \emph{production systems}. Compiled PyTorch, FlexAttention, and \ours run under \texttt{torch.compile}, FA4 and FlashInfer under their own JIT compilers.

\paragraph{Correctness}
To ensure exactness, we validate every output and gradient tensor against an independent FP32 reference, computed eagerly from the relation equations by gathering the relation vector for each query--key pair. An implementation passes if, when cast to FP32, \emph{each} of its output and gradient tensors is elementwise within 5\% relative and 0.03 absolute error, or has a relative $\ell_2$ error within eight rounding units of the target data type (0.78\% for FP16). We mark systems that fail this check with \textdagger{} in Table~\ref{tab:systems}. The full results, presented in \cref{app:h200,app:gpus}, mark them with \textdagger{} as well. \Ours \emph{always} passes this check.

\paragraph{Hardware and software}
Unless noted otherwise, we run all experiments on a single NVIDIA H200 SXM GPU using CUDA~13.0, PyTorch~2.14.1, Triton~3.8.0, FA4~4.0.0b33, FlashInfer~0.7.2, and FlashDeBERTa~0.0.7.

\subsection{Attention-Level Performance}
\label{sec:eval-attention}
We benchmark one attention layer, from $Q$, $K$, $V$ and the relation tables to $O$ and, backward, to their gradients.

\paragraph{Setup}
Following FlashAttention-2~\citep{flashattention2}, we fix a budget of 16K tokens and sweep $N$ from 128 to 8K tokens with batch size $B=16\text{K}/N$, $H=12$ heads of dimension $d=64$, FP16, and bidirectional attention. We time forward passes and backward passes (with stored relevant forward pass values) and report throughput as useful FLOPs over latency, crediting every system the same work, so throughput ratios are latency ratios. Peak memory is recorded as the device memory allocated beyond the inputs. \Cref{app:experimental-details} details the FLOP count and the timing protocol.

\paragraph{Throughput}
\Cref{fig:kernels} compares \ours with the fastest system of each kind. \Ours outperforms every production system at every length except \speedupLosses{}, where our setup overhead dominates. For Shaw, DeBERTa, and T5, its lead increases as sequences grow, to \speedupForward{} in the forward and \speedupBackward{} in the backward pass at 8K tokens. This is expected, since every production baseline either reads a dense $N\times N$ bias or obtains its relation terms score by score, while \ours instead reads one entry per query and per key wherever $|\delta|\ge\Delta$, so its share of individual lookups vanishes as $N$ grows. TXL gains less (1.4--2.4$\times$), as its identity map gives every offset its own row and keeps the projections quadratic in $N$ for all systems, including \ours. We note that for DeBERTa's backward pass, every production system fails our FP32 check from 4K tokens (\cref{tab:systems}), while \ours remains exact. Full results are in \Cref{tab:throughput-h200} in \cref{app:h200}.

\paragraph{Memory footprint}
\Cref{tab:memory-h200} compares the peak memory of \ours with eager PyTorch, the leanest production baseline, and the leanest engine running our decomposition at 128 and 8K tokens, with full results in \cref{tab:memory-lengths-h200} in \cref{app:h200}. At 128 tokens, \ours is on par with the leanest system, needing the least memory or within a few tens of MiB of it. In terms of memory, systems that build the dense bias keep growing with $N$, and at 8K tokens, \ours needs \memoryForward{} less forward and \memoryBackward{} less backward memory than the leanest production system. Every system's footprint grows monotonically with $N$, so these findings stay consistent with sequence length. The one exception is DeBERTa's forward pass, where the inference-only GLiNER4j recomputes the relation products in every tile instead of storing projections. However, as shown in \cref{fig:kernels}, this comes at a significant throughput cost, where \ours outperforms it $3.7\times$.

\paragraph{Is the relation decomposition enough?}
Our decomposition (\cref{eq:projected-relation-scores}) also fits the per-score callbacks of programmable engines. We thus compare integrating only this component against \ours as a whole. In the forward pass, this lifts the engines above their dense-bias form at long sequences. \Ours still stays \decompositionForward{} faster at 8K tokens (\cref{tab:decomposition}). In the backward pass, the gap widens to \decompositionBackward{}, because the engines accumulate the table gradients with an atomic addition per score. Atomic contention also costs memory: on every mechanism but T5, the engines' backward pass needs at least as much as the backward pass of \ours.

\paragraph{Different GPU architectures}
On Blackwell and Ampere architectures, namely RTX PRO 6000 and RTX A6000, \ours generalizes well, as it keeps its lead over the production systems at 8K tokens (\cref{tab:gpus}), with the same pattern as on the H200: the largest gains on Shaw, and the smallest on TXL. We provide the full results in \cref{app:gpus}.

\begin{table}[t]
\centering\footnotesize
\setlength{\tabcolsep}{2.3pt}
\caption{Improvements of \ours at 8K tokens. \emph{Top:} against implementations with our decomposition (H200). \emph{Bottom:} speedup ($\times$) over the fastest production system. ``F''/``B'': forward/backward.}
\label{tab:decomposition}
\label{tab:gpus}
\vspace{0pt}
\begin{tabular}{@{}lrrrrrrrr@{}}
\toprule
 & \multicolumn{2}{c}{Shaw} & \multicolumn{2}{c}{TXL} & \multicolumn{2}{c}{DeBERTa} & \multicolumn{2}{c}{T5} \\
\cmidrule(l{2pt}r{2pt}){2-3} \cmidrule(l{2pt}r{2pt}){4-5} \cmidrule(l{2pt}r{2pt}){6-7} \cmidrule(l{2pt}r{2pt}){8-9}
 & F & B & F & B & F & B & F & B \\
\midrule
\multicolumn{9}{@{}l}{\textit{vs.\ our decomposition (H200)}} \\
Speedup & 1.3 & 3.5 & 1.1 & 1.7 & 2.2 & 4.3 & 1.5 & 228 \\
Memory saving & 0.99 & 2.9 & 1.0 & 3.0 & 0.51 & 2.2 & 1.0 & 0.75 \\
\midrule
\multicolumn{9}{@{}l}{\textit{vs.\ production systems: speedup per GPU}} \\
H200 (9.0) & 6.4 & 21 & 1.8 & 2.0 & 3.7 & 7.1 & 4.6 & 13 \\
RTX PRO 6000 (12.0) & 15 & 19 & 2.0 & 2.3 & 3.2 & 8.4 & 6.4 & 12 \\
RTX A6000 (8.6) & 12 & 25 & 2.1 & 2.1 & 2.7 & 6.5 & 5.1 & 7.4 \\
\bottomrule
\end{tabular}
\end{table}

\begin{table}[t]
\centering\footnotesize
\setlength{\tabcolsep}{2.3pt}
\caption{Peak memory (MiB) on the H200 (FP16) at 128 and 8K.}
\label{tab:memory-h200}
\vspace{0pt}
\begin{tabular}{@{}c@{\hspace{5pt}}lrrrrrrrr@{}}
\toprule
 & & \multicolumn{2}{c}{Shaw} & \multicolumn{2}{c}{TXL} & \multicolumn{2}{c}{DeBERTa} & \multicolumn{2}{c}{T5} \\
\cmidrule(l{2pt}r{2pt}){3-4} \cmidrule(l{2pt}r{2pt}){5-6} \cmidrule(l{2pt}r{2pt}){7-8} \cmidrule(l{2pt}r{2pt}){9-10}
 & MiB & 128 & 8K & 128 & 8K & 128 & 8K & 128 & 8K \\
\midrule
\multirow{4}{*}{\rotatebox[origin=c]{90}{\textit{Forward}}} & PyTorch eager & 192 & 14K & 288 & 18K & 216 & 12K & 144 & 7.5K \\
 & Leanest baseline & \textbf{98} & 14K & 168 & 9.0K & \textbf{48} & \textbf{48} & \textbf{24} & 3.0K \\
\cmidrule{2-10}
 & + our decomp. & 120 & \textbf{221} & \textbf{120} & \textbf{3.1K} & 216 & 408 & \textbf{24} & \textbf{24} \\
 & \textbf{\ourslabel} & 120 & 222 & \textbf{120} & \textbf{3.1K} & 216 & 804 & \textbf{24} & \textbf{24} \\
\midrule
\multirow{4}{*}{\rotatebox[origin=c]{90}{\textit{Backward}}} & PyTorch eager & 168 & 19K & 264 & 12K & 288 & 9.0K & 168 & 9.0K \\
 & Leanest baseline & 144 & 11K & 264 & 9.1K & 288 & 648 & 74 & 5.1K \\
\cmidrule{2-10}
 & + our decomp. & 360 & 666 & 360 & 9.2K & 552 & 1.0K & \textbf{73} & \textbf{73} \\
 & \textbf{\ourslabel} & \textbf{129} & \textbf{229} & \textbf{129} & \textbf{3.1K} & \textbf{177} & \textbf{470} & 97 & 97 \\
\bottomrule
\end{tabular}
\end{table}

\subsection{End-to-End Performance}
\label{sec:eval-e2e}

To show practical efficiency, we apply \ours to three released models, targeting speech transcription and long-audio fine-tuning with FastConformer's TXL attention, and long-document classification with DeBERTa-v3's disentangled attention. All experiments were run on one H200 in BF16 mixed precision, with the software and recipes in \cref{app:experimental-details}. We report the median time, excluding the one-time compilation and autotuning of \ours, which takes under 3 minutes and is cached afterwards. For speech, we also report the word error rate (WER) and the inverse real-time factor (RTFx)~\citep{rtfx}, the seconds of audio transcribed per wall-clock second.

\paragraph{Transcription with Parakeet-TDT-0.6B-v3}
Parakeet-TDT-0.6B-v3 \citep{canary} pairs a FastConformer encoder with a token-and-duration transducer (TDT) decoder~\citep{xu2023tdt}, and \ours replaces only the encoder's attention. We transcribe the TED-LIUM~3 test set \citep{hernandez2018tedlium3} in NeMo \citep{kuchaiev2019nemo}, once as short segments and once as whole talks of up to 25 minutes (\cref{tab:parakeet}). We compare against NeMo's production attention paths, native and SDPA~(\cref{app:experimental-details}). \Ours matches the WER of native attention, making it a drop-in replacement for trained models. Since the rest of the model runs unchanged, its gain is bounded by the share of time spent in attention, unlike in the isolated layer (\cref{sec:eval-attention}). On short segments, this share is small, and \ours is \parakeetSegmentsSpeedupSDPA{} faster than SDPA. On whole talks, attention dominates, and \ours nearly doubles throughput while needing a fifth of the memory of native attention. In fine-tuning, the transducer loss adds an overhead that dominates each step, so \ours brings no speedup at any length.

\begin{table}[t]
\centering\footnotesize
\setlength{\tabcolsep}{4pt}
\caption{Parakeet-TDT-0.6B-v3 transcription of TED-LIUM~3's test set in NeMo, H200, BF16, as segments and as whole talks.}
\label{tab:parakeet}
\vspace{0pt}
\begin{tabular}{@{}lrrrrrr@{}}
\toprule
 & \multicolumn{3}{c}{Segments} & \multicolumn{3}{c}{Whole talks} \\
\cmidrule(l{2pt}r{2pt}){2-4} \cmidrule(l{2pt}){5-7}
Attention & WER & RTFx & GiB & WER & RTFx & GiB \\
\midrule
NeMo (native) & 2.80 & 2726 & \textbf{4.7} & 2.88 & 492 & 53 \\
NeMo (SDPA) & 2.82 & 3761 & \textbf{4.7} & 3.01 & 613 & 27 \\
\midrule
\textbf{\ourslabel} & 2.80 & \textbf{4560} & \textbf{4.7} & 2.89 & \textbf{1103} & \textbf{11} \\
\bottomrule
\end{tabular}
\end{table}

\paragraph{Fine-tuning Canary-1B-v2 on long audio}
Canary-1B-v2 \citep{canary} pairs a FastConformer encoder with a Transformer autoregressive decoder. Due to the autoregressive decoder, \ours brings little or no speedup in token generation, and helps only during prefill, i.e., the encoder pass, which takes at most a quarter of the inference time, and under 5\% from 5-minute windows on. To test training, we fine-tune all its parameters on TED-LIUM~3 talks cut into windows of 30 seconds to 18 minutes, and test on windows of the same length (\cref{tab:canary}). Training losses match NeMo's SDPA path at every window both run, and test WER agrees up to 5-minute windows, beyond which a few long recordings dominate the differences (\cref{app:experimental-details}). The gain again follows attention's share of each step, so \ours speeds up training steps by \canarySpeedup{}, more for longer windows. SDPA runs out of memory from 15-minute windows, while \ours fits both by recomputing its projections in the backward pass, scaling to longer contexts.

\begin{table}[t]
\centering\footnotesize
\setlength{\tabcolsep}{2.5pt}
\caption{Fine-tuning Canary-1B-v2 on windows of up to $W$. \emph{oom} denotes when training ran out of memory.}
\label{tab:canary}
\vspace{0pt}
\begin{tabular}{@{}lllllll@{}}
\toprule
 & \multicolumn{2}{l}{Step (ms)} & \multicolumn{2}{l}{Peak (GiB)} & \multicolumn{2}{l}{WER (\%)} \\
\cmidrule(r{4pt}){2-3} \cmidrule(r{4pt}){4-5} \cmidrule{6-7}
$W$ & SDPA & \textbf{\ourslabel} & SDPA & \textbf{\ourslabel} & SDPA & \textbf{\ourslabel} \\
\midrule
30 s & 281 & \textbf{257} (1.09$\times$) & 44 & 43 & 3.73 & 3.82 \\
2 min & 388 & \textbf{325} (1.19$\times$) & 49 & 49 & 3.20 & 3.04 \\
5 min & 506 & \textbf{390} (1.30$\times$) & 60 & 60 & 3.35 & 3.24 \\
9 min & 740 & \textbf{546} (1.36$\times$) & 79 & 79 & 3.73 & 4.34 \\
15 min & \emph{oom} & 876 & \emph{oom} & 99 & \emph{oom} & 4.44 \\
18 min & \emph{oom} & 1082 & \emph{oom} & 129 & \emph{oom} & 6.40 \\
\bottomrule
\end{tabular}
\end{table}

\paragraph{Long-document classification with DeBERTa-v3}
Finally, for text, following \citet{condevaux2023lsg}, we fine-tune DeBERTa-v3-base to classify BIGPATENT \citep{sharma2019bigpatent} patents into nine CPC sections, with inputs truncated to 512, 2K, or 4K tokens, and compare with a Transformers pipeline (\cref{tab:deberta}). Both reach the same test accuracy within \debertaAccuracyGap{} points, and longer inputs improve it. \Ours more than doubles training speed from 2K tokens on. Its peak memory also stops growing at 2K tokens, while Transformers' reaches almost four times as much at 4K.

\providecommand{\pending}{\textcolor{red}{TBD}}
\begin{table}[t]
\centering\footnotesize
\setlength{\tabcolsep}{2.5pt}
\caption{Fine-tuning DeBERTa-v3-base with content truncated to $L$ tokens. HF denotes Transformers' performance.}
\label{tab:deberta}
\vspace{0pt}
\begin{tabular}{@{}lllllll@{}}
\toprule
 & \multicolumn{2}{l}{Accuracy (\%)} & \multicolumn{2}{l}{Step (ms)} & \multicolumn{2}{l}{Peak (GiB)} \\
\cmidrule(r{4pt}){2-3} \cmidrule(r{4pt}){4-5} \cmidrule{6-7}
$L$ & HF & \textbf{\ourslabel} & HF & \textbf{\ourslabel} & HF & \textbf{\ourslabel} \\
\midrule
512 & 66.6 & 66.7 & 116 & \textbf{105} (1.10$\times$) & 8.2 & \textbf{6.1} \\
2K & 67.3 & 67.6 & 942 & \textbf{430} (2.19$\times$) & 15 & \textbf{7.1} \\
4K & 67.7 & 67.6 & 2185 & \textbf{915} (2.39$\times$) & 27 & \textbf{7.1} \\
\bottomrule
\end{tabular}
\end{table}

\subsection{Ablation Studies}
\label{sec:eval-ablations}
To isolate the efficacy of each part of our contribution, \cref{tab:ablations} switches off one component of \ours at a time, and registers the memory and efficiency tradeoff. \Cref{app:ablations} further reports the results on every mechanism and the HBM traffic.

Two methods carry most of the speedup. Without atomic-free gradients, the backward pass is \ablationAccumulationLatencyBackward{} slower, needs \ablationAccumulationMemoryBackward{} the memory, and loses determinism. The decomposition engines use atomics, hence their backward gap in \cref{sec:eval-attention}. Without projections, every tile computes its own relation products, which slows the forward pass by \ablationProjectionLatencyForward{} and the backward pass even more, but needs only \ablationProjectionMemoryForward{} the memory, the same tradeoff GLiNER4j makes.

The other methods are more targeted, with considerable gains of up to \ablationTargetedLatency{} in time or \ablationTargetedMemory{} in memory on the maps they serve (\cref{tab:ablations-full}). Reachable ranges save memory, mainly on TXL's identity map, while constant-range reuse speeds up the clipped and bucketed maps. Bucket copies add further speed at a memory cost. Finally, recomputing the projections instead of retaining them cuts memory to \ablationRetentionMemoryBackward{} for a \ablationRetentionLatencyBackward{} slowdown, which is how \ours is able to fit Canary's longest windows (\cref{sec:eval-e2e}).

\begin{table}[t]
\centering\footnotesize
\caption{Ablations on an H200 (FP16): latency and peak memory without each method, relative to \ours, as the range over the mechanisms it applies to. ``--'': not applicable.}
\label{tab:ablations}
\vspace{0pt}
\setlength{\tabcolsep}{3pt}
\begin{tabular}{@{}lcccc@{}}
\toprule
w/o & \multicolumn{2}{c}{Latency ($\times$)} & \multicolumn{2}{c}{Peak memory ($\times$)} \\
\cmidrule(lr){2-3} \cmidrule(l){4-5}
(128--8K) & Fwd & Bwd & Fwd & Bwd \\
\midrule
Atomic-free gradients & -- & 2.9--4.4 & -- & 1.8--2.6 \\
Projections & 2.0--2.9 & 3.5--5.4 & 0.19--0.43 & 0.20--0.43 \\
Reachable ranges & 1.0--1.1 & 1.0--1.1 & 1.1--1.5 & 1.1--1.5 \\
Constant-range reuse & 1.1--1.3 & 1.2--1.3 & 1.0 & 1.0 \\
Bucket copies & 1.0--1.1 & 1.1 & 0.70--1.0 & 0.81--1.0 \\
Retained projections & -- & 1.1 & -- & 0.42--0.50 \\
\bottomrule
\end{tabular}
\end{table}

\vspace{-0.6em}
\section{Limitations}
\label{sec:limitations}

While \ours resolves the main bottlenecks of relative-position attention, several limitations remain. First, identity maps such as Transformer-XL's assign a distinct relation row to every offset, so their projection work and space remain quadratic in sequence length. Computing the projections inside each tile instead makes the space linear but is \ablationProjectionLatencyForward{} slower in the forward pass (\cref{tab:ablations}). Second, maps whose relation rows do not cover consecutive offsets fall back to FP32 atomic accumulation, which is slower and not deterministic, although every evaluated mechanism avoids it. Third, \ours favors generality across mechanisms, maps, and GPUs over hardware specialization. It autotunes its tile sizes but uses no architecture-specific primitives such as warp specialization or asynchronous bulk transfers~\citep{flashattention3,flashattention4}, which could yield further speedups and which we leave as future work.

\vspace{-0.6em}
\section{Conclusion}
\label{sec:conclusion}
We presented an exact tiled algorithm that unifies and accelerates additive relative-position attention. By restricting relation projections to reachable ranges, reusing entries across constant-offset tiles, and accumulating gradients by offset without atomics, \ours removes the dense bias of existing backends, with its memory and bandwidth costs, and keeps gradients deterministic for identity, clipped, and bucketed maps. Across mechanisms and GPUs, \ours is a drop-in replacement for long-context training and inference in production models, so these mechanisms no longer need to be substituted for efficiency.

\section*{Acknowledgments}
Parts of this work were done at INSAIT, Sofia University ``St. Kliment Ohridski'', Bulgaria. This work was partially funded by the Ministry of Education and Science of Bulgaria (support for INSAIT, part of the Bulgarian National Roadmap for Research Infrastructure). This project was supported with computational resources provided by Google Cloud Platform (GCP).

\bibliography{references}
\bibliographystyle{mlsys2025}

\clearpage
\appendix
\crefalias{section}{appendix}
\crefalias{subsection}{subappendix}
\crefalias{subsubsection}{subsubappendix}
\section{Notation}
\label{app:notation}
\Cref{tab:notation} lists the symbols of \cref{sec:background,sec:relaccel} and \cref{app:derivations}. We consider one sequence and one attention head, and all indices are zero-based. Symbols local to a single proof are defined where they are used.

\begin{table*}[t]
\centering\footnotesize
\caption{Notation.}
\label{tab:notation}
\vspace{5pt}
\begin{tabular}{@{}l>{\raggedright\arraybackslash}p{0.78\textwidth}@{}}
\toprule
Symbol & Meaning \\
\midrule
\multicolumn{2}{@{}l}{\textit{Tiled Attention}} \\
$N,d$ & Sequence length and head dimension. \\
$B,H$ & Batch size and number of heads. \\
$i,j$ & Query and key positions in $\{0,\ldots,N-1\}$. \\
$Q,K,V,O$ & Query, key, value, and output matrices in $\mathbb R^{N\times d}$, with rows $q_i^\top,k_j^\top,v_j^\top,o_i^\top$. \\
$S,P$ & Complete softmax logits and row-normalized probabilities in $\mathbb R^{N\times N}$. \\
$\tau,M$ & Normalization scalar and additive attention mask, $M_{ij}\in\{0,-\infty\}$. \\
$\mathcal I,\mathcal J$ & Query and key tiles: sets of consecutive query and key positions processed together on chip. \\
$m_i,\ell_i,u_i$ & Running row maximum, denominator sum, and unnormalized output of the online softmax. \\
$L_i,D_i$ & Row log-normalizer $L_i=m_i+\log\ell_i$, and $D_i=\langle dO_i,o_i\rangle$. \\
$dX,\mathbf 1$ & Gradient of the training loss with respect to $X$, and the all-ones vector. \\
\midrule
\multicolumn{2}{@{}l}{\textit{Relative-Position Attention}} \\
$\phi(i,j),\bar\phi$ & Discrete map assigning pair $(i,j)$ its relation-row index. For offset-based maps, $\phi(i,j)=\bar\phi(i-j)$. \\
$r,R$ & Relation-row index and number of referenced rows. \\
$\delta,\Delta$ & Offset $\delta=i-j$ and saturation limit $\Delta$, beyond which all offsets map to row $0$ or $R-1$. \\
$\operatorname{clip}(\delta,-\Delta,\Delta)$ & The offset $\delta$ clamped to $[-\Delta,\Delta]$. \\
$A^Q,A^K,b$ & Query- and key-relation vector tables in $\mathbb R^{R\times d}$, with rows $(a^Q_r)^\top,(a^K_r)^\top$, and scalar-bias table $b\in\mathbb R^R$. \\
$Z^Q,Z^K$ & Relation projections in $\mathbb R^{N\times R}$, $Z^Q_{ir}=q_i^\top a^Q_r$ and $Z^K_{jr}=k_j^\top a^K_r$. \\
\midrule
\multicolumn{2}{@{}l}{\Ours} \\
$G,g$ & Group size and group index. \\
$Q_g,K_g$ & Group $g$ of $G$ contiguous queries and keys, positions $gG,\ldots,gG+G-1$. \\
$\Omega^Q_g,\Omega^K_g$ & Relation rows reached by $Q_g$ and $K_g$, at most $\min(N+G-1,R)$ each, and $A^Q_{\Omega^Q_g}$ holds those rows of $A^Q$. For identity maps, $\Omega^Q_g=\{gG,\ldots,gG+N+G-2\}$ and $\Omega^K_g=\{N-gG-G,\ldots,2N-gG-2\}$. \\
$s,\beta(s),E$ & Bucket copies: copied row $s\in\{0,\ldots,2\Delta\}$, its bucket $\beta(s)=\bar\phi(s-\Delta)$, and copy matrix $E\in\{0,1\}^{(2\Delta+1)\times R}$ with $E_{sr}=[\beta(s)=r]$. \\
$\tilde\phi$ & Clipped map that replaces a bucketed $\phi$ after copying. \\
$\tilde Z^Q,\tilde Z^K$ & Per-offset projections, $\tilde Z^Q=Z^QE^\top$ for bucketed maps and $Z^Q$ otherwise. \\
$d\tilde Z^Q,d\tilde Z^K$ & Gradient entries accumulated by offset, folded into $dZ^Q,dZ^K$. \\
\bottomrule
\end{tabular}
\end{table*}

\section{Gradient Derivations and Tiling Correctness}
\label{app:derivations}
This appendix derives the gradients of relative-position attention (\cref{app:derivations-gradients}) and shows that the tiled passes of \ours compute them exactly (\cref{app:derivations-tiling}). All claims hold in exact arithmetic; in floating point, the tiled passes only fix the order of the same sums.

\paragraph{Setup}
We consider one head and write $\mathcal{L}$ for the training loss and $dX=\partial\mathcal{L}/\partial X$. By \cref{eq:token-relation-projections}, the canonical scores (\cref{eq:canonical-score}) equal
\begin{equation}
  \label{eq:app-scores}
  S_{ij} = \tau\bigl(q_i^\top k_j + Z^Q_{ir} + Z^K_{jr} + b_r\bigr) + M_{ij},
\end{equation}
with $r=\phi(i,j)$, $P=\operatorname{softmax}_{\text{row}}(S)$, and $O=PV$. Absent terms are zero. We write $[\cdot]$ for the indicator of a condition and assume each query has at least one unmasked key. For offset-based maps we write $\phi(i,j)=\bar\phi(i-j)$, where $\bar\phi$ is nondecreasing and saturates at $\Delta$ (\cref{sec:background-relative}).

\subsection{Gradients}
\label{app:derivations-gradients}

\begin{lemma}[Softmax backward]
\label{lem:app-softmax}
Let $dP_{ij}=\langle dO_i,v_j\rangle$ and $D_i=\langle dO_i,o_i\rangle$. Then
\begin{equation*}
  dV = P^\top dO, \qquad dS_{ij} = P_{ij}\,(dP_{ij}-D_i).
\end{equation*}
\end{lemma}
\begin{proof}
\textit{Step 1.} Since $o_i=\sum_j P_{ij}v_j$ is linear in $P$ and $V$, we get $dP_{ij}=\langle dO_i,v_j\rangle$ and $dv_j=\sum_i P_{ij}\,dO_i$, that is, $dV=P^\top dO$.

\textit{Step 2.} Row $i$ of $P$ depends only on row $i$ of $S$, with $\partial P_{it}/\partial S_{ij}=P_{it}([t=j]-P_{ij})$.

\textit{Step 3.} By the chain rule,
\begin{align*}
  dS_{ij}&=\textstyle\sum_t dP_{it}P_{it}([t=j]-P_{ij})\\
  &=P_{ij}\bigl(dP_{ij}-\textstyle\sum_t P_{it}\,dP_{it}\bigr).
\end{align*}

\textit{Step 4.} By Step 1, $\sum_t P_{it}\,dP_{it}=\langle dO_i,\sum_t P_{it}v_t\rangle=\langle dO_i,o_i\rangle=D_i$.
\end{proof}
Masked pairs have $P_{ij}=0$ and thus $dS_{ij}=0$, so they contribute nothing to any gradient below.

\begin{lemma}[Score gradients]
\label{lem:app-scores}
Treating $Q,K,Z^Q,Z^K,b$ in \cref{eq:app-scores} as independent inputs, their gradients are $\tau\,dS\,K$ for $Q$, $\tau\,dS^\top Q$ for $K$, and \cref{eq:projection-gradients} for $Z^Q,Z^K,b$.
\end{lemma}
\begin{proof}
\textit{Step 1.} $S_{ij}$ is affine in each input, with partial derivatives
\begin{align*}
  \partial S_{ij}/\partial q_{i'} &= \tau[i'=i]\,k_j,\\
  \partial S_{ij}/\partial Z^Q_{i'r} &= \tau[i'=i]\,[\phi(i,j)=r],\\
  \partial S_{ij}/\partial Z^K_{j'r} &= \tau[j'=j]\,[\phi(i,j)=r],\\
  \partial S_{ij}/\partial b_r &= \tau[\phi(i,j)=r],
\end{align*}
and symmetrically for $k_{j'}$.

\textit{Step 2.} Each gradient is $\sum_{i,j}dS_{ij}$ times the matching partial derivative. The indicators $[i'=i]$ and $[j'=j]$ remove one sum, and $[\phi(i,j)=r]$ restricts the other to the pairs mapped to row $r$, which gives \cref{eq:projection-gradients}.
\end{proof}
Each pair $(i,j)$ thus adds $\tau\,dS_{ij}$ to exactly one entry of each of $dZ^Q$, $dZ^K$, and $db$, namely row $\phi(i,j)$. We define $dZ^Q$ and $dZ^K$ by \cref{eq:projection-gradients} even when $A^Q$ or $A^K$ is absent.

\begin{lemma}[Input and table gradients]
\label{lem:app-inputs}
With $Z^Q=Q(A^Q)^\top$ and $Z^K=K(A^K)^\top$,
\begin{align*}
  dQ &= \tau\,dS\,K + dZ^QA^Q, & dA^Q &= (dZ^Q)^\top Q,\\
  dK &= \tau\,dS^\top Q + dZ^KA^K, & dA^K &= (dZ^K)^\top K,
\end{align*}
$db=(dZ^Q)^\top\mathbf 1=(dZ^K)^\top\mathbf 1$, and $dV=P^\top dO$.
\end{lemma}
\begin{proof}
\textit{Step 1.} $Z^Q_{ir}=q_i^\top a^Q_r$ is bilinear, so $\partial Z^Q_{ir}/\partial q_i=a^Q_r$ and $\partial Z^Q_{ir}/\partial a^Q_r=q_i$.

\textit{Step 2.} Query $q_i$ reaches $\mathcal L$ along two paths, directly through $S_{i\cdot}$ and through row $i$ of $Z^Q$. The chain rule adds both, giving $dq_i=\tau\sum_j dS_{ij}k_j+\sum_r dZ^Q_{ir}a^Q_r$, the $i$-th row of $dQ$.

\textit{Step 3.} Row $a^Q_r$ reaches $\mathcal L$ only through column $r$ of $Z^Q$, so $da^Q_r=\sum_i dZ^Q_{ir}q_i$, the $r$-th row of $dA^Q$.

\textit{Step 4.} Summing \cref{eq:projection-gradients} over $i$ gives $\sum_i dZ^Q_{ir}=\tau\sum_{i,j:\,\phi(i,j)=r}dS_{ij}=db_r$, and likewise over $j$ for $dZ^K$.

\textit{Step 5.} The relation terms enter only $S$, so $dV$ is unchanged from \cref{lem:app-softmax}. Keys follow by symmetry.
\end{proof}

\begin{lemma}[Bucket expansion]
\label{lem:app-buckets}
Let $\phi$ be bucketed with $R$ buckets and saturation limit $\Delta$. Let $\beta(s)=\bar\phi(s-\Delta)$ for $s\in\{0,\ldots,2\Delta\}$, and let $E\in\{0,1\}^{(2\Delta+1)\times R}$ have entries $E_{sr}=[\beta(s)=r]$. Replacing $(A^Q,A^K,b)$ by $(EA^Q,EA^K,Eb)$ and $\phi$ by the clipped map $\tilde\phi$ leaves every score unchanged, and the gradients of the original tables are
\begin{equation*}
  dZ^Q=d\tilde Z^QE,\qquad dZ^K=d\tilde Z^KE,\qquad db=E^\top d\tilde b.
\end{equation*}
\end{lemma}
\begin{proof}
\textit{Step 1.} For $|\delta|<\Delta$, $\tilde\phi$ sends $\delta$ to row $s=\delta+\Delta$, and $\beta(s)=\bar\phi(\delta)$. For $\delta\ge\Delta$, $\tilde\phi$ sends $\delta$ to $s=2\Delta$, and $\beta(2\Delta)=\bar\phi(\Delta)=R-1=\bar\phi(\delta)$ by saturation, and likewise for $\delta\le-\Delta$. Hence $\beta(\tilde\phi(i,j))=\phi(i,j)$.

\textit{Step 2.} Row $s$ of $EA^Q$ is $a^Q_{\beta(s)}$, so $\tilde Z^Q=Q(EA^Q)^\top=Z^QE^\top$, and $\tilde Z^Q_{i\tilde\phi(i,j)}=Z^Q_{i\phi(i,j)}$ by Step 1. The same holds for $\tilde Z^K$ and $Eb$, so every score is unchanged.

\textit{Step 3.} The map $Z^Q\mapsto Z^QE^\top$ is linear, so its backward is $dZ^Q=d\tilde Z^QE$, and likewise for $Z^K$ and $b$. Entrywise, each bucket's gradient is the sum of its copies, $dZ^Q_{ir}=\sum_{s:\,\beta(s)=r}d\tilde Z^Q_{is}$.
\end{proof}
After this fold, \cref{lem:app-inputs} gives the gradients of $A^Q$, $A^K$, and $b$ (lines~\ref{line:bwd-fold}--\ref{line:bwd-reduce} of \cref{alg:backward}).

\subsection{Tiling}
\label{app:derivations-tiling}
Let $\mathcal I$ and $\mathcal J$ be tiles of consecutive queries and keys.

\begin{lemma}[Tile locality]
\label{lem:app-locality}
The score tile $S_{\mathcal I\mathcal J}$ depends only on $Q_{\mathcal I}$, $K_{\mathcal J}$, $M_{\mathcal I\mathcal J}$, rows $\mathcal I$ of $Z^Q$, rows $\mathcal J$ of $Z^K$, and $b$. Given $L_{\mathcal I}$ and $D_{\mathcal I}$, the tiles $P_{\mathcal I\mathcal J}$ and $dS_{\mathcal I\mathcal J}$ further depend only on $S_{\mathcal I\mathcal J}$, $V_{\mathcal J}$, and $dO_{\mathcal I}$.
\end{lemma}
\begin{proof}
\textit{Step 1.} In \cref{eq:app-scores}, $S_{ij}$ reads only $q_i$, $k_j$, $M_{ij}$, row $i$ of $Z^Q$, row $j$ of $Z^K$, and $b$.

\textit{Step 2.} With $L_i=\log\sum_t e^{S_{it}}$, we have $P_{ij}=e^{S_{ij}-L_i}$. Combined with \cref{lem:app-softmax}, this gives \cref{eq:prob-reconstruction}, which reads only the listed tiles.
\end{proof}

\begin{lemma}[Online softmax]
\label{lem:app-online}
Let $U_t=\mathcal J_1\cup\cdots\cup\mathcal J_t$ be the keys processed after $t$ iterations of the inner loop of \cref{alg:forward}. Then, for each $i\in\mathcal I$,
\begin{gather*}
  m_i=\max_{j\in U_t}S_{ij},\qquad
  \ell_i=\sum_{j\in U_t}e^{S_{ij}-m_i},\\
  u_i=\sum_{j\in U_t}e^{S_{ij}-m_i}v_j.
\end{gather*}
After the last key tile, $o_i=u_i/\ell_i$ and $L_i=\log\sum_j e^{S_{ij}}$.
\end{lemma}
\begin{proof}
\textit{Step 1 (base).} For $t=0$, $U_0=\emptyset$, and the initial values $-\infty,0,0$ are the empty maximum and sums.

\textit{Step 2 (induction).} Assume the claim for $t-1$. The update \cref{eq:online-softmax} gives $m_i'=\max(\max_{U_{t-1}}S_{ij},\max_{\mathcal J_t}S_{ij})=\max_{U_t}S_{ij}$, and
\begin{align*}
  \ell_i'&=e^{m_i-m_i'}\!\!\sum_{j\in U_{t-1}}\!\!e^{S_{ij}-m_i}+\sum_{j\in\mathcal J_t}e^{S_{ij}-m_i'}\\
  &=\sum_{j\in U_t}e^{S_{ij}-m_i'},
\end{align*}
and the same computation holds for $u_i'$.

\textit{Step 3 (output).} After all tiles, $U_t$ covers every key, so $u_i/\ell_i=\sum_j e^{S_{ij}}v_j/\sum_j e^{S_{ij}}=\sum_j P_{ij}v_j=o_i$ and $m_i+\log\ell_i=\log\sum_je^{S_{ij}}$.
\end{proof}
By \cref{lem:app-locality}, each iteration can add the relation terms to $S_{\mathcal I\mathcal J}$ on chip before the update, so \cref{lem:app-online} applies to the scores of \cref{eq:app-scores} unchanged, and \cref{alg:forward} returns the exact $O$ and $L$.

\begin{lemma}[Saturated tiles]
\label{lem:app-saturation}
Let $\delta_{\min}=\min\mathcal I-\max\mathcal J$ and $\delta_{\max}=\max\mathcal I-\min\mathcal J$, and let $\phi$ be offset-based.
\begin{enumerate}[label=(\alph*),nosep]
  \item Every pair $(i,j)\in\mathcal I\times\mathcal J$ has $\delta_{\min}\le i-j\le\delta_{\max}$.
  \item If $\delta_{\min}\ge\Delta$, every pair has $\phi(i,j)=R-1$. If $\delta_{\max}\le-\Delta$, every pair has $\phi(i,j)=0$.
  \item In either case, with that row $r$, the relation terms of the tile are $\tau\bigl(z^Q\mathbf 1^\top+\mathbf 1(z^K)^\top+b_r\mathbf 1\mathbf 1^\top\bigr)$, where $z^Q_i=Z^Q_{ir}$ and $z^K_j=Z^K_{jr}$.
\end{enumerate}
\end{lemma}
\begin{proof}
\textit{Step 1.} For $i\in\mathcal I$ and $j\in\mathcal J$, $i\ge\min\mathcal I$ and $j\le\max\mathcal J$, so $i-j\ge\delta_{\min}$. The upper bound follows in the same way.

\textit{Step 2.} By saturation, $\bar\phi(\delta)=R-1$ for all $\delta\ge\Delta$ and $\bar\phi(\delta)=0$ for all $\delta\le-\Delta$. Step 1 places every offset of the tile on one of these sides.

\textit{Step 3.} Substituting the constant $r$ into \cref{eq:app-scores} gives (c).
\end{proof}
A saturated tile thus reads $|\mathcal I|+|\mathcal J|+1$ entries instead of $3|\mathcal I||\mathcal J|$ (line~\ref{line:saturated} of \cref{alg:forward}). The test is sufficient but not necessary, since a tile that fails it may still contain saturated pairs. Both branches assign every pair its row $\phi(i,j)$, so the test changes only the cost, not the result.

\begin{lemma}[Reachable ranges]
\label{lem:app-ranges}
Let query group $g$ hold queries $i\in\{gG,\ldots,gG+G-1\}$, and let all $N$ keys be visible.
\begin{enumerate}[label=(\alph*),nosep]
  \item The group reaches the $N+G-1$ consecutive offsets $\delta\in\{gG-N+1,\ldots,gG+G-1\}$.
  \item These offsets map to at most $\min(N+G-1,R)$ consecutive rows $\Omega^Q_g$. For identity maps, $\Omega^Q_g=\{gG,\ldots,gG+N+G-2\}$ and $\Omega^K_g=\{N-gG-G,\ldots,2N-gG-2\}$.
  \item If a query tile $\mathcal I$ lies inside group $g$, every entry $Z^Q_{i\phi(i,j)}$ that its score tiles read is a column of $Q_g(A^Q_{\Omega^Q_g})^\top$.
\end{enumerate}
Key groups and $\Omega^K_g$ follow by symmetry.
\end{lemma}
\begin{proof}
\textit{Step 1.} With $i\in[gG,gG+G-1]$ and $j\in[0,N-1]$, the offset $\delta=i-j$ ranges from $gG-(N-1)$ to $gG+G-1$, and every value in between occurs.

\textit{Step 2.} A nondecreasing $\bar\phi$ maps an interval of offsets onto an interval of rows of at most the same size, and there are only $R$ rows. For identity maps, $r=\delta+N-1$ shifts the interval of Step 1.

\textit{Step 3.} Every pair read by the tile has $i$ in group $g$ and thus $\phi(i,j)\in\Omega^Q_g$ by Step 2.
\end{proof}
With $G=128$ equal to the largest tile size and tiles aligned to groups, every query tile lies inside one group, as (c) requires. For clipped and bucketed maps after expansion, $R=2\Delta+1$, so each query stores at most $\min(N+G-1,2\Delta+1)$ entries.

\begin{proposition}[Atomic-free accumulation]
\label{prop:app-atomic-free}
Let $\phi$ be offset-based, and let each block of the query-owned pass of \cref{alg:backward} own one query tile $\mathcal I$.
\begin{enumerate}[label=(\alph*),nosep]
  \item For fixed $i$, $j\mapsto\delta=i-j$ is a bijection from the keys onto $\{i-N+1,\ldots,i\}$, so
  \begin{equation*}
    dZ^Q_{ir}=\tau\!\!\sum_{\delta:\,\bar\phi(\delta)=r}\!\!dS_{i,i-\delta}.
  \end{equation*}
  \item Each entry $dZ^Q_{ir}$ with $i\in\mathcal I$ is written by a single block, without atomic additions, and its summation order is fixed.
  \item For $\delta\ge\Delta$ and for $\delta\le-\Delta$, the sum in (a) collapses into one boundary sum each, for rows $R-1$ and $0$.
\end{enumerate}
The key-owned pass and $dZ^K$ follow by symmetry, using $i\mapsto\delta=i-j$ for fixed $j$.
\end{proposition}
\begin{proof}
\textit{Step 1.} The map $j\mapsto i-j$ is injective, and its image over $j\in\{0,\ldots,N-1\}$ is $\{i-N+1,\ldots,i\}$. Reindexing the sum over $j$ in \cref{eq:projection-gradients} by $\delta$ gives (a).

\textit{Step 2.} By (a), row $i$ of $dZ^Q$ depends only on row $i$ of $dS$. By \cref{lem:app-locality}, the block that owns $\mathcal I$ can compute these rows tile by tile, splitting the sum in (a) over key tiles. No other block touches row $i$, so the block keeps each entry on chip and writes it once. Within the block, each pair adds to the single entry $(i,\bar\phi(i-j))$ in the order of the key-tile loop, so the order is fixed.

\textit{Step 3.} By saturation, all $\delta\ge\Delta$ share row $R-1$ and all $\delta\le-\Delta$ share row $0$, so their terms in (a) form one sum per row.
\end{proof}

\begin{corollary}[Exact backward pass]
\label{cor:app-backward}
\Cref{alg:backward} returns the exact gradients $dQ,dK,dV,dA^Q,dA^K,db$ of \cref{lem:app-inputs}.
\end{corollary}
\begin{proof}
\textit{Step 1.} By \cref{lem:app-locality}, each block rebuilds $S_{\mathcal I\mathcal J}$, $P_{\mathcal I\mathcal J}$, and $dS_{\mathcal I\mathcal J}$ exactly from the saved $L$ and $D$.

\textit{Step 2.} Every sum over keys, in $\tau\,dS\,K$ and $dZ^Q$, splits over key tiles and is completed by the query-owned pass (\cref{prop:app-atomic-free}). Every sum over queries, in $P^\top dO$, $\tau\,dS^\top Q$, and $dZ^K$, splits over query tiles and is completed by the key-owned pass.

\textit{Step 3.} For bucketed maps, \cref{lem:app-buckets} folds the copies into $dZ^Q$ and $dZ^K$. The final products are then exactly those of \cref{lem:app-inputs}.
\end{proof}

\section{Kernel and Algorithm Details}
\label{app:kernel-details}
\Cref{sec:implementation} summarizes our implementation. Here, we detail how \ours maps \cref{alg:forward,alg:backward} onto kernels, how it reduces the relation gradients deterministically, how it stores relation projections in half precision, and which inputs and tile sizes its kernels support.

\paragraph{Kernel structure}
\Cref{tab:kernels} lists the kernels of \ours in launch order, with the lines of \cref{alg:forward,alg:backward} that each one runs. The forward pass first projects every query group and key group onto its reachable range and stores the preprojected relation entries in HBM. The attention kernel then runs the tiled loop and writes $O$ and $L$. The backward pass computes $D$ and runs the query-parallel pass, whose gradient entries are reduced into $dQ$, $dA^Q$, and $db$ before the key-parallel pass starts. Both passes can therefore write their gradient entries into the same buffer. The key-parallel pass and its reduction then produce $dK$, $dV$, and $dA^K$. If the forward pass does not save $Z^Q$ and $Z^K$, the backward pass recomputes them with the projection kernel before it starts. When both $A^Q$ and $b$ are present, the projection kernel adds $b$ into the stored $Z^Q$, so each score reads two entries instead of three. Bucket copies (line~\ref{line:copy}) run before the projection and copy the rows of the tables rather than the columns of the projections, $\tilde A^Q=EA^Q$, which gives the same $\tilde Z^Q=Q(\tilde A^Q)^\top=Z^QE^\top$.

\begin{table}[t]
\centering\footnotesize
\setlength{\tabcolsep}{2.5pt}
\caption{Kernels of \ours in launch order, parallelized over heads.}
\label{tab:kernels}
\vspace{5pt}
\begin{tabular}{@{}llll@{}}
\toprule
Kernel & Lines & Parallel over & Writes \\
\midrule
\multicolumn{4}{@{}l}{\textit{Forward (\cref{alg:forward})}} \\
Projection & \ref{line:project} & tokens, relation rows & $Z^Q,Z^K$ \\
Attention & \ref{line:fwd-loop}--\ref{line:fwd-write} & queries & $O,L$ \\
\midrule
\multicolumn{4}{@{}l}{\textit{Backward (\cref{alg:backward})}} \\
Preprocessing & \ref{line:bwd-delta} & queries & $D$ \\
Query-parallel pass & \ref{line:bwd-qloop}--\ref{line:bwd-qwrite} & queries & $dQ,d\tilde Z^Q$ \\
Query reduction & \ref{line:bwd-qreduce} & tokens, relation rows & $dQ,dA^Q,db$ \\
Key-parallel pass & \ref{line:bwd-kloop}--\ref{line:bwd-kwrite} & keys & $dK,dV,d\tilde Z^K$ \\
Key reduction & \ref{line:bwd-reduce} & tokens, relation rows & $dK,dA^K$ \\
\bottomrule
\end{tabular}
\end{table}

\paragraph{Deterministic table gradients}
Each table gradient is a sum over all tokens. \Ours splits the tokens into chunks $c$ and computes $dA^Q=\sum_c (dZ^Q_c)^\top Q_c$, and likewise $dA^K$ and $db$, in two kernels. In the first, each thread block computes the term of one chunk and writes it to HBM in FP32. In the second, these terms are summed in a fixed order. For bucketed maps, the gradients of the copied rows are then summed back into their buckets, $dA^Q=E^\top d\tilde A^Q$, also in a fixed order. No gradient is accumulated atomically, so the backward pass is deterministic.

\paragraph{Atomic fallback}
Accumulating gradients by offset (\cref{sec:relaccel-gradients}) requires each relation row to cover one contiguous interval of offsets. This holds for identity, clipped, and bucketed maps, and \ours checks it once per map on the host. Maps that fail it fall back to kernels that accumulate their relation gradients with FP32 atomic additions. Examples are maps that give $\delta$ and $-\delta$ the same row, or maps that depend on more than the offset.

\paragraph{Half-precision numerics}
Inputs may be in FP16, BF16, or FP32. The online softmax uses base-2 exponentials, with $\log_2 e$ folded into $\tau$, and its statistics, the accumulators, and the boundary sums stay in FP32 on chip. In HBM, the gradient entries use the input precision. The relation projections use FP16 for both FP16 and BF16 inputs. BF16 gives up three mantissa bits for a wider range, so FP16 stores the projections with an eight times smaller rounding error. For BF16 inputs, the projections are clamped to the FP16 range of $\pm65504$, so that an outlier cannot overflow to infinity.

\paragraph{Masks, packing, and dropout}
The kernels support padding masks, causal masks, and packed variable-length sequences, in which each thread block processes a query tile of one sequence. Attention dropout draws its mask from a counter-based random number generator inside the kernels, so the backward pass regenerates the mask of the forward pass instead of storing it.

\paragraph{Specialization and tuning}
The sequence-length buckets end at 64, 128, 384, 512, 768, 1K, 2K, 4K, and 8K tokens, and longer sequences use the 8K bucket. Besides the specializations of \cref{sec:implementation}, each kernel is compiled separately for lengths that are and are not a multiple of 16, so a bucket needs at most two compilations per kernel. For multiples of 16, Triton can use wider memory accesses, so our model adapters pad each sequence to a multiple of 16 tokens and mask the padding. For the same reason, each group's range of projection columns starts and ends on a multiple of 16. Our lookup tables cover compute capabilities 8.0, 8.6, and 9.0 (A100, RTX A6000, and H100 or H200), with one entry per kernel, precision, head dimension, bucket, and set of relation terms. The hardware resource heuristic starts from a $64\times64$ tile and shrinks it until its estimated registers and shared memory fit the GPU. Instead of both, \ours can autotune each kernel at run time over a short list of candidate tile sizes, once per bucket. All measurements in \cref{sec:evaluation} use autotuning, which the end-to-end runs complete in the precompilation step before timing. The exception is the ablation that computes the projections inside each tile (\cref{tab:ablations}), which uses its fastest fixed tile sizes.

\section{Additional Experimental Details}
\label{app:experimental-details}

\paragraph{Kernel benchmarks}
A forward pass is credited $4BHN^2d$ FLOPs for its two attention GEMMs plus $2BHNRd$ for each relation projection over $R$ table rows: one for Shaw and TXL ($A^Q$), two for DeBERTa ($A^Q$ and $A^K$), and none for T5. A backward pass is credited twice as much, for its four attention GEMMs and the two gradient GEMMs of each projection. FlashAttention-2~\citep{flashattention2} instead credits 2.5 times the forward FLOPs, counting the recomputed scores, which we leave out because not every system recomputes them. Each measurement runs in a fresh process with \texttt{triton.testing.do\_bench}, which flushes the L2 cache before each sample. After 50ms of warmup, we report the median of at least ten timed runs. Compilation and autotuning happen before timing and are excluded. Systems under \texttt{torch.compile} use its \texttt{max-autotune} mode with static shapes and without CUDA graphs. Within each mechanism and length, the systems run in a random order, and we record GPU clocks, power, and temperature instead of locking the clocks. The RTX PRO 6000 and RTX A6000 use the same software as the H200.

\paragraph{End-to-end software}
The speech models run in NeMo~3.0.0 on PyTorch~2.9.1 and CUDA~12.8, and DeBERTa on the software stack of \cref{sec:eval-setup}.

\paragraph{Parakeet transcription}
The test set has 1,155 segments and 11 talks of up to ${\sim}25$ minutes, which we transcribe one at a time. We report word error rate (WER) after Whisper's text normalization \citep{radford2023whisper}, for which NVIDIA's model card reports 2.75\% \citep{nvidia2025parakeetv3}, and inverse real-time factor (RTFx), the audio seconds processed per wall-clock second. NeMo's FlexAttention path captures sequence-length-dependent index arithmetic and bias tensors in its \texttt{score\_mod} closure, so each new input length triggers a \texttt{torch.compile} recompilation that erases any speedup, and running it uncompiled falls back to a much slower unfused implementation.

\paragraph{Canary fine-tuning}
We fine-tune on 100 hours of TED-LIUM~3 training talks, cut into windows of at most 30 seconds and 2, 5, 9, 15, and 18 minutes, for 3 epochs with about 540 seconds of audio per step. Final-epoch training losses match within \canaryLossGap{} at every window both attentions run. As Whisper does, we decode transcripts that fall into repetition loops again in shorter pieces. From 9-minute windows on, such loops affect one or two long recordings per window, and these recordings account for the remaining WER differences.

\paragraph{DeBERTa fine-tuning}
The patent data has 25,000 training, 5,000 validation, and 5,000 test documents of 3,900 tokens on average. We train for 3 epochs with AdamW~\citep{loshchilov2019adamw} at a learning rate of $2\times10^{-5}$, 10\% linear warmup, weight decay 0.01, and 16 patents per step in 4K-token micro-batches, and report test accuracy at the best validation epoch. Macro-F1 agrees within \debertaFOneGapRest{} points from 2K tokens and differs by \debertaFOneGapShort{} points at 512 tokens. \citet{condevaux2023lsg} report 66.6 micro-F1 and 61.8 macro-F1 for RoBERTa on this task.

\section{Additional Ablations}
\label{app:ablations}
\Cref{tab:ablations-full} gives every method of \cref{tab:ablations} for each mechanism and pass.

\begin{table*}[t]
\centering
\footnotesize
\caption{Ablations without each method, relative to \ours: (a) latency and peak memory on an H200 (FP16), as the geometric mean over 128 to 8K tokens, and (b) HBM traffic (bytes read and written) on an H100 at 4K tokens. Retained projections span 4 layers. Without projections, each tile computes its own relation products, timed at its fastest tile configuration. --: not applicable.}
\label{tab:ablations-full}
\label{tab:hbm}
\vspace{5pt}
\textit{(a) Latency and peak memory (H200)}\par\vspace{3pt}
\begin{tabular}{@{}llrrrrrrrr@{}}
\toprule
 & & \multicolumn{2}{c}{Shaw} & \multicolumn{2}{c}{TXL} & \multicolumn{2}{c}{DeBERTa} & \multicolumn{2}{c}{T5} \\
\cmidrule(l{2pt}r{2pt}){3-4} \cmidrule(l{2pt}r{2pt}){5-6} \cmidrule(l{2pt}r{2pt}){7-8} \cmidrule(l{2pt}r{2pt}){9-10}
w/o & Pass & Lat. & Mem. & Lat. & Mem. & Lat. & Mem. & Lat. & Mem. \\
\midrule
Atomic-free gradients & Bwd & 3.92 & 1.75 & 2.94 & 2.60 & 3.99 & 2.16 & 4.39 & 1.97 \\
Projections & Fwd & 2.47 & 0.43 & 1.99 & 0.20 & 2.91 & 0.19 & -- & -- \\
 & Bwd & 4.98 & 0.43 & 3.48 & 0.20 & 5.41 & 0.22 & -- & -- \\
Reachable ranges & Fwd & -- & -- & 1.12 & 1.48 & 1.01 & 1.08 & -- & -- \\
 & Bwd & -- & -- & 1.09 & 1.50 & 1.01 & 1.08 & -- & -- \\
Constant-range reuse & Fwd & 1.11 & 1.00 & -- & -- & 1.23 & 1.00 & 1.26 & 1.00 \\
 & Bwd & 1.20 & 1.00 & -- & -- & 1.18 & 1.00 & 1.29 & 1.00 \\
Bucket copies & Fwd & -- & -- & -- & -- & 1.04 & 0.70 & 1.13 & 1.00 \\
 & Bwd & -- & -- & -- & -- & 1.10 & 0.81 & 1.09 & 1.00 \\
Retained projections & Bwd & -- & -- & 1.11 & 0.50 & 1.10 & 0.42 & -- & -- \\
\bottomrule
\end{tabular}
\par\vspace{10pt}
\textit{(b) HBM traffic (H100)}\par\vspace{3pt}
\begin{tabular}{@{}llrrrr@{}}
\toprule
w/o & Pass & Shaw & TXL & DeBERTa & T5 \\
\midrule
Atomic-free gradients & Bwd & 1.6 & 3.1 & 1.8 & 1.3 \\
Reachable ranges & Fwd & -- & 1.4 & 1.0 & -- \\
 & Bwd & -- & 1.4 & 1.0 & -- \\
Constant-range reuse & Fwd & 0.98 & -- & 0.97 & 1.0 \\
 & Bwd & 1.0 & -- & 0.95 & 1.0 \\
Bucket copies & Fwd & -- & -- & 0.55 & 1.1 \\
 & Bwd & -- & -- & 0.86 & 1.0 \\
\bottomrule
\end{tabular}
\end{table*}

\paragraph{HBM traffic}
We also count the bytes each pass reads from and writes to HBM with Nsight Compute, on an H100 at 4K tokens, with each method switched off (\cref{tab:hbm}). Atomic accumulation adds the traffic of full FP32 gradient buffers to the backward pass, \trafficAccumulation{} that of \ours and \trafficAccumulationTXL{} on TXL, whose identity map has the most entries. Without reachable ranges, TXL stores and rereads projections for every offset instead of only the reachable ones and moves \trafficRowGroupsTXL{} the data, while DeBERTa, whose groups already reach nearly all rows, stays at \trafficRowGroupsDeBERTa{}. Constant-range reuse leaves the traffic unchanged at \trafficConstantRanges{}: without it, a saturated tile rereads the same entry for every score, and these repeated reads hit in on-chip caches, so its speedup comes from fewer loads and index computations rather than from fewer bytes. Bucket copies store a projection column per offset instead of per bucket, which raises DeBERTa's traffic by \trafficExpansionDeBERTa{}. Computing the relation products inside each tile reads no stored projections, so Shaw and TXL move only \trafficProjection{} the data, but every tile recomputes them.

\section{Detailed Results on the H200}
\label{app:h200}
\Cref{tab:throughput-h200,tab:memory-lengths-h200} give the throughput and peak memory of every system behind \cref{fig:kernels} and \cref{tab:memory-h200} at every sequence length. FlexAttention, FA4, and FlashInfer without our decomposition run on the model's dense bias.

\paragraph{Memory across lengths}
The footprint of \ours stops growing beyond 1K tokens for Shaw, DeBERTa, and T5, and grows only with the projections for TXL (\cref{tab:memory-lengths-h200}). GLiNER4j's lower forward memory on DeBERTa comes from recomputing the relation products in every tile instead of storing projections. The projections ablation (\cref{tab:ablations}) shows this trade within \ours: less memory, but a \ablationProjectionLatencyForward{} slower forward pass.

\begin{table*}[t]
\centering
\footnotesize
\setlength{\tabcolsep}{3pt}
\caption{Throughput of every system on the H200 (FP16) at every length. Engines running our decomposition use their faster variant per mechanism and pass. \textbf{Bold}: the fastest system. \textdagger: fails the FP32 check. --: not supported.}
\label{tab:throughput-h200}
\vspace{5pt}
\begin{tabular}{@{}llrrrrrrr@{\hspace{10pt}}rrrrrrr@{}}
\toprule
 & & \multicolumn{7}{c}{Forward (TFLOP/s $\uparrow$)} & \multicolumn{7}{c}{Backward (TFLOP/s $\uparrow$)} \\
\cmidrule(lr){3-9} \cmidrule(l){10-16}
 & System & 128 & 256 & 512 & 1K & 2K & 4K & 8K & 128 & 256 & 512 & 1K & 2K & 4K & 8K \\
\midrule
Shaw & PyTorch eager & 27 & 37 & 31 & 26 & 18 & 12 & 7.1 & 45 & 64 & 57 & 41 & 27 & 15 & 8.0 \\
 & PyTorch compile & 72 & 89 & 84 & 82 & 62 & 57 & 44 & 64 & 89 & 19 & 4.6 & 1.7 & 0.74 & 0.35 \\
 & PyTorch SDPA & 67 & 89 & 68 & 49 & 30 & 17 & 8.9 & 67 & 74 & 60 & 41 & 26 & 15 & 7.8 \\
 & FlexAttention & 54 & 66 & 51 & 37 & 25 & 15 & 8.3 & 61 & 78 & 64 & 44 & 27 & 15 & 7.9 \\
 & FA4 & 45 & 53 & 45 & 33 & 23 & 14 & 8.1 & -- & -- & -- & -- & -- & -- & -- \\
 & FlashInfer & 46 & 50 & 40 & 29 & 21 & 13 & 7.8 & -- & -- & -- & -- & -- & -- & -- \\
\cmidrule{2-16}
 & FlexAttention + decomp. & 90 & 130 & 112 & 121 & 129 & 135 & 138 & 56 & 65 & 60 & 56 & 52 & 49 & 47 \\
 & FA4 + decomp. & 79 & 104 & 116 & 137 & 176 & 197 & 211 & -- & -- & -- & -- & -- & -- & -- \\
 & FlashInfer + decomp. & 89 & 108 & 113 & 124 & 148 & 155 & 160 & -- & -- & -- & -- & -- & -- & -- \\
 & Flashlight + decomp. & 71 & 80 & 85 & 84 & 84 & 85 & 86 & -- & -- & -- & -- & -- & -- & -- \\
 & \textbf{\ourslabel} & \textbf{117} & \textbf{157} & \textbf{144} & \textbf{177} & \textbf{216} & \textbf{249} & \textbf{279} & \textbf{77} & \textbf{96} & \textbf{96} & \textbf{113} & \textbf{134} & \textbf{153} & \textbf{166} \\
\midrule
TXL & PyTorch eager & 12 & 14 & 16 & 17 & 18 & 17 & 17 & 21 & 24 & 27 & 28 & 29 & 30 & 30 \\
 & PyTorch compile & 54 & 76 & 92 & 76 & 91 & 93 & 79 & 30 & 43 & 56 & 66 & 72 & 75 & 66 \\
 & PyTorch SDPA & 15 & 17 & 19 & 20 & 21 & 21 & 22 & 24 & 25 & 27 & 28 & 29 & 29 & 29 \\
 & FlexAttention & 15 & 18 & 20 & 21 & 21 & 21 & 22 & 23 & 25 & 28 & 29 & 29 & 29 & 29 \\
 & FA4 & 14 & 16 & 19 & 20 & 21 & 21 & 21 & -- & -- & -- & -- & -- & -- & -- \\
 & FlashInfer & 15 & 16 & 18 & 19 & 20 & 20 & 20 & -- & -- & -- & -- & -- & -- & -- \\
 & NeMo FlexAttention & 68 & 100 & 122 & 133 & 139 & 140 & 134 & 30 & 42 & 52 & 57 & 60 & 61 & 54 \\
\cmidrule{2-16}
 & FlexAttention + decomp. & 85 & 125 & 146 & 168 & 182 & 189 & 194 & 52 & 61 & 67 & 75 & 78 & 79 & 79 \\
 & FA4 + decomp. & 76 & 101 & 135 & 156 & 185 & 196 & 202 & -- & -- & -- & -- & -- & -- & -- \\
 & FlashInfer + decomp. & 85 & 104 & 141 & 161 & 190 & 203 & 208 & -- & -- & -- & -- & -- & -- & -- \\
 & NeMo FlexAttention + decomp. & 85 & 125 & 145 & 168 & 181 & 189 & 193 & 52 & 61 & 67 & 75 & 78 & 79 & 78 \\
 & Flashlight + decomp. & 38 & 54 & 68 & 80 & 88 & 89 & 90 & -- & -- & -- & -- & -- & -- & -- \\
 & \textbf{\ourslabel} & \textbf{113} & \textbf{151} & \textbf{176} & \textbf{204} & \textbf{220} & \textbf{228} & \textbf{236} & \textbf{74} & \textbf{92} & \textbf{108} & \textbf{119} & \textbf{128} & \textbf{132} & \textbf{133} \\
\midrule
DeBERTa & PyTorch eager & 16 & 17 & 17 & 16 & 13 & 11 & 10 & 25 & 27 & 28 & 24 & 19 & 16\rlap{\textsuperscript{\textdagger}} & 14\rlap{\textsuperscript{\textdagger}} \\
 & PyTorch compile & 58 & 58 & 70 & 51 & 46 & 43 & 37 & 29 & 24 & 18 & 2.8 & 5.9 & 5.1\rlap{\textsuperscript{\textdagger}} & 5.0\rlap{\textsuperscript{\textdagger}} \\
 & FlexAttention & 20 & 21 & 21 & 20 & 16 & 14 & 13 & 28 & 29 & 29 & 25 & 19 & 16\rlap{\textsuperscript{\textdagger}} & 14\rlap{\textsuperscript{\textdagger}} \\
 & FA4 & 19 & 19 & 20 & 19 & 15 & 13 & 13 & -- & -- & -- & -- & -- & -- & -- \\
 & FlashInfer & 19 & 19 & 19 & 18 & 14 & 13 & 12 & -- & -- & -- & -- & -- & -- & -- \\
 & FlashDeBERTa & 21 & 28 & 33 & 33 & 33 & 33 & 33 & 12 & 11 & 7.9 & 2.6 & 1.4 & 1.1\rlap{\textsuperscript{\textdagger}} & 1.0\rlap{\textsuperscript{\textdagger}} \\
 & GLiNER4j & 87 & 82 & 70 & 54 & 47 & 44 & 43 & -- & -- & -- & -- & -- & -- & -- \\
\cmidrule{2-16}
 & FlexAttention + decomp. & 84 & \textbf{89} & \textbf{82} & 71 & 72 & 73 & 74 & 38 & 34 & 36 & 33 & 28 & 25 & 24 \\
 & FA4 + decomp. & 82 & 65 & 58 & 49 & 52 & 52 & 51 & -- & -- & -- & -- & -- & -- & -- \\
 & FlashInfer + decomp. & 75 & 66 & 77 & 66 & 68 & 65 & 63 & -- & -- & -- & -- & -- & -- & -- \\
 & Flashlight + decomp. & 66 & 64 & 60 & 57 & 56 & 56 & 56 & -- & -- & -- & -- & -- & -- & -- \\
 & \textbf{\ourslabel} & \textbf{92} & 88 & \textbf{82} & \textbf{80} & \textbf{101} & \textbf{129} & \textbf{159} & \textbf{65} & \textbf{64} & \textbf{57} & \textbf{55} & \textbf{66} & \textbf{84} & \textbf{101} \\
\midrule
T5 & PyTorch eager & 16 & 23 & 30 & 33 & 22 & 20 & 18 & 27 & 39 & 45 & 42 & 34 & 22 & 12 \\
 & PyTorch compile & 45 & 53 & 70 & 86 & 60 & 54 & 67 & 33 & 33 & 19 & 8.0 & 3.5 & 1.7 & 0.80 \\
 & PyTorch SDPA & 6.7 & 8.9 & 11 & 12 & 11 & 11 & 9.9 & 13 & 18 & 20 & 21 & 18 & 14 & 9.2 \\
 & FlexAttention & 66 & 123 & 141 & 176 & 120 & 77 & 45 & \textbf{60} & \textbf{81} & 81 & 66 & 43 & 24 & 13 \\
 & FA4 & 45 & 65 & 100 & 104 & 82 & 60 & 39 & -- & -- & -- & -- & -- & -- & -- \\
 & FlashInfer & 58 & 71 & 112 & 116 & 94 & 65 & 40 & -- & -- & -- & -- & -- & -- & -- \\
\cmidrule{2-16}
 & FlexAttention + decomp. & 86 & 113 & 133 & 144 & 150 & 153 & 155 & 7.4 & 1.7 & 1.1 & 0.94 & 0.85 & 0.81 & 0.76 \\
 & FA4 + decomp. & 66 & 87 & 134 & 153 & 184 & 195 & 203 & -- & -- & -- & -- & -- & -- & -- \\
 & FlashInfer + decomp. & 77 & 86 & 123 & 132 & 153 & 155 & 158 & -- & -- & -- & -- & -- & -- & -- \\
 & Flashlight + decomp. & 79 & 95 & 103 & 108 & 110 & 112 & 114 & -- & -- & -- & -- & -- & -- & -- \\
 & \textbf{\ourslabel} & \textbf{105} & \textbf{143} & \textbf{191} & \textbf{238} & \textbf{266} & \textbf{288} & \textbf{307} & 45 & 67 & \textbf{95} & \textbf{121} & \textbf{144} & \textbf{162} & \textbf{174} \\
\bottomrule
\end{tabular}
\end{table*}

\begin{table*}[t]
\centering
\footnotesize
\setlength{\tabcolsep}{3pt}
\caption{Peak memory of every system on the H200 (FP16) at every length. Engines running our decomposition use their faster variant per mechanism and pass. \textbf{Bold}: the leanest system. \textdagger: fails the FP32 check. --: not supported.}
\label{tab:memory-lengths-h200}
\vspace{5pt}
\begin{tabular}{@{}llrrrrrrr@{\hspace{10pt}}rrrrrrr@{}}
\toprule
 & & \multicolumn{7}{c}{Forward (MiB $\downarrow$)} & \multicolumn{7}{c}{Backward (MiB $\downarrow$)} \\
\cmidrule(lr){3-9} \cmidrule(l){10-16}
 & System & 128 & 256 & 512 & 1K & 2K & 4K & 8K & 128 & 256 & 512 & 1K & 2K & 4K & 8K \\
\midrule
Shaw & PyTorch eager & 192 & 336 & 624 & 1.2K & 2.3K & 5.1K & 14K & 168 & 312 & 600 & 1.1K & 2.3K & 4.7K & 19K \\
 & PyTorch compile & 122 & 224 & 440 & 920 & 2.0K & 5.0K & 14K & 144 & 193 & 289 & 512 & 1.3K & 3.6K & 11K \\
 & PyTorch SDPA & \textbf{98} & 200 & 418 & 904 & 2.0K & 5.1K & 14K & 169 & 264 & 456 & 840 & 1.6K & 4.7K & 19K \\
 & FlexAttention & \textbf{98} & 200 & 418 & 904 & 2.0K & 5.1K & 14K & 216 & 360 & 648 & 1.2K & 2.3K & 4.7K & 19K \\
 & FA4 & \textbf{98} & 200 & 418 & 904 & 2.0K & 5.1K & 14K & -- & -- & -- & -- & -- & -- & -- \\
 & FlashInfer & \textbf{98} & 200 & 418 & 904 & 2.0K & 5.1K & 14K & -- & -- & -- & -- & -- & -- & -- \\
\cmidrule{2-16}
 & FlexAttention + decomp. & 122 & 176 & 224 & 224 & 224 & 224 & 224 & 360 & 522 & 666 & 666 & 666 & 666 & 666 \\
 & FA4 + decomp. & 120 & \textbf{174} & 222 & 222 & 222 & 222 & 222 & -- & -- & -- & -- & -- & -- & -- \\
 & FlashInfer + decomp. & 120 & \textbf{174} & 222 & 222 & 222 & 222 & 222 & -- & -- & -- & -- & -- & -- & -- \\
 & Flashlight + decomp. & 120 & 217 & \textbf{221} & \textbf{221} & \textbf{221} & \textbf{221} & \textbf{221} & -- & -- & -- & -- & -- & -- & -- \\
 & \textbf{\ourslabel} & 120 & \textbf{174} & 222 & 222 & 222 & 222 & 222 & \textbf{129} & \textbf{181} & \textbf{229} & \textbf{229} & \textbf{229} & \textbf{229} & \textbf{229} \\
\midrule
TXL & PyTorch eager & 288 & 576 & 1.1K & 2.3K & 4.5K & 9.1K & 18K & 264 & 456 & 840 & 1.6K & 3.1K & 6.1K & 12K \\
 & PyTorch compile & 216 & 360 & 648 & 1.2K & 2.3K & 4.6K & 9.1K & 312 & 408 & 696 & 1.2K & 2.4K & 4.6K & 9.1K \\
 & PyTorch SDPA & 288 & 576 & 1.1K & 2.3K & 4.5K & 9.1K & 18K & 264 & 456 & 840 & 1.6K & 3.1K & 6.1K & 12K \\
 & FlexAttention & 288 & 576 & 1.1K & 2.3K & 4.5K & 9.1K & 18K & 264 & 456 & 840 & 1.6K & 3.1K & 6.1K & 12K \\
 & FA4 & 288 & 576 & 1.1K & 2.3K & 4.5K & 9.1K & 18K & -- & -- & -- & -- & -- & -- & -- \\
 & FlashInfer & 288 & 576 & 1.1K & 2.3K & 4.5K & 9.1K & 18K & -- & -- & -- & -- & -- & -- & -- \\
 & NeMo FlexAttention & 168 & 288 & 576 & 1.1K & 2.2K & 4.5K & 9.0K & 264 & 408 & 696 & 1.2K & 2.4K & 4.6K & 9.1K \\
\cmidrule{2-16}
 & FlexAttention + decomp. & 122 & 176 & 272 & 464 & 848 & \textbf{1.6K} & \textbf{3.1K} & 360 & 522 & 810 & 1.4K & 2.5K & 4.7K & 9.2K \\
 & FA4 + decomp. & \textbf{120} & \textbf{174} & \textbf{270} & \textbf{462} & \textbf{846} & \textbf{1.6K} & \textbf{3.1K} & -- & -- & -- & -- & -- & -- & -- \\
 & FlashInfer + decomp. & \textbf{120} & \textbf{174} & \textbf{270} & \textbf{462} & \textbf{846} & \textbf{1.6K} & \textbf{3.1K} & -- & -- & -- & -- & -- & -- & -- \\
 & NeMo FlexAttention + decomp. & 122 & 176 & 272 & 464 & 848 & \textbf{1.6K} & \textbf{3.1K} & 360 & 522 & 810 & 1.4K & 2.5K & 4.7K & 9.2K \\
 & Flashlight + decomp. & 181 & 278 & 472 & 859 & 1.6K & 3.1K & 6.1K & -- & -- & -- & -- & -- & -- & -- \\
 & \textbf{\ourslabel} & \textbf{120} & \textbf{174} & \textbf{270} & \textbf{462} & \textbf{846} & \textbf{1.6K} & \textbf{3.1K} & \textbf{129} & \textbf{181} & \textbf{277} & \textbf{469} & \textbf{859} & \textbf{1.6K} & \textbf{3.1K} \\
\midrule
DeBERTa & PyTorch eager & 216 & 384 & 770 & 1.5K & 3.0K & 6.1K & 12K & 288 & 372 & 600 & 1.1K & 2.3K & 4.5K\rlap{\textsuperscript{\textdagger}} & 9.0K\rlap{\textsuperscript{\textdagger}} \\
 & PyTorch compile & 312 & 453 & 624 & 804 & 1.2K & 1.9K & 3.4K & 408 & 621 & 793 & 794 & 1.1K & 1.9K\rlap{\textsuperscript{\textdagger}} & 3.4K\rlap{\textsuperscript{\textdagger}} \\
 & FlexAttention & 216 & 384 & 770 & 1.5K & 3.0K & 6.1K & 12K & 288 & 372 & 648 & 1.2K & 2.3K & 4.6K\rlap{\textsuperscript{\textdagger}} & 9.1K\rlap{\textsuperscript{\textdagger}} \\
 & FA4 & 216 & 384 & 770 & 1.5K & 3.0K & 6.1K & 12K & -- & -- & -- & -- & -- & -- & -- \\
 & FlashInfer & 216 & 384 & 770 & 1.5K & 3.0K & 6.1K & 12K & -- & -- & -- & -- & -- & -- & -- \\
 & FlashDeBERTa & 768 & 768 & 768 & 768 & 768 & 768 & 768 & 648 & 648 & 648 & 648 & 648 & 648\rlap{\textsuperscript{\textdagger}} & 648\rlap{\textsuperscript{\textdagger}} \\
 & GLiNER4j & \textbf{48} & \textbf{48} & \textbf{48} & \textbf{48} & \textbf{48} & \textbf{48} & \textbf{48} & -- & -- & -- & -- & -- & -- & -- \\
\cmidrule{2-16}
 & FlexAttention + decomp. & 218 & 326 & 518 & 806 & 806 & 806 & 806 & 552 & 822 & 1.0K & 1.0K & 1.0K & 1.0K & 1.0K \\
 & FA4 + decomp. & 216 & 324 & 516 & 804 & 804 & 804 & 804 & -- & -- & -- & -- & -- & -- & -- \\
 & FlashInfer + decomp. & 216 & 324 & 408 & 408 & 408 & 408 & 408 & -- & -- & -- & -- & -- & -- & -- \\
 & Flashlight + decomp. & 216 & 409 & 789 & 802 & 802 & 802 & 802 & -- & -- & -- & -- & -- & -- & -- \\
 & \textbf{\ourslabel} & 216 & 324 & 516 & 804 & 804 & 804 & 804 & \textbf{177} & \textbf{229} & \textbf{325} & \textbf{470} & \textbf{470} & \textbf{470} & \textbf{470} \\
\midrule
T5 & PyTorch eager & 144 & 242 & 438 & 840 & 1.6K & 3.4K & 7.5K & 168 & 312 & 600 & 1.1K & 2.3K & 4.5K & 9.0K \\
 & PyTorch compile & 96 & 144 & 240 & 432 & 816 & 1.5K & 3.0K & 120 & 193 & 385 & 769 & 1.6K & 3.1K & 6.0K \\
 & PyTorch SDPA & 432 & 626 & 1.0K & 1.9K & 3.7K & 7.3K & 15K & 336 & 624 & 1.2K & 2.3K & 4.5K & 9.0K & 18K \\
 & FlexAttention & 26 & 27 & 32 & 50 & 192 & 768 & 3.0K & 74 & 77 & 92 & 152 & 392 & 1.3K & 5.1K \\
 & FA4 & \textbf{24} & 26 & 30 & 48 & 192 & 768 & 3.0K & -- & -- & -- & -- & -- & -- & -- \\
 & FlashInfer & \textbf{24} & 26 & 30 & 48 & 192 & 768 & 3.0K & -- & -- & -- & -- & -- & -- & -- \\
\cmidrule{2-16}
 & FlexAttention + decomp. & 26 & 26 & 26 & 26 & 26 & 26 & 26 & \textbf{73} & \textbf{73} & \textbf{73} & \textbf{73} & \textbf{73} & \textbf{73} & \textbf{73} \\
 & FA4 + decomp. & \textbf{24} & \textbf{24} & \textbf{24} & \textbf{24} & \textbf{24} & \textbf{24} & \textbf{24} & -- & -- & -- & -- & -- & -- & -- \\
 & FlashInfer + decomp. & \textbf{24} & \textbf{24} & \textbf{24} & \textbf{24} & \textbf{24} & \textbf{24} & \textbf{24} & -- & -- & -- & -- & -- & -- & -- \\
 & Flashlight + decomp. & \textbf{24} & \textbf{24} & \textbf{24} & \textbf{24} & \textbf{24} & \textbf{24} & \textbf{24} & -- & -- & -- & -- & -- & -- & -- \\
 & \textbf{\ourslabel} & \textbf{24} & \textbf{24} & \textbf{24} & \textbf{24} & \textbf{24} & \textbf{24} & \textbf{24} & 97 & 97 & 97 & 97 & 97 & 97 & 97 \\
\bottomrule
\end{tabular}
\end{table*}

\section{Results on Other GPUs}
\label{app:gpus}
\Cref{tab:throughput-rtx_pro_6000_blackwell_server_edition,tab:throughput-rtx_a6000} give the throughput of every system on the RTX PRO 6000 Blackwell and the RTX A6000 at every sequence length, as \cref{tab:throughput-h200} does on the H200.

\begin{table*}[t]
\centering
\footnotesize
\setlength{\tabcolsep}{3pt}
\caption{Throughput of every system on the RTX PRO 6000 Blackwell (FP16) at every length. Engines running our decomposition use their faster variant per mechanism and pass. \textbf{Bold}: the fastest system. \textdagger: fails the FP32 check. --: not supported.}
\label{tab:throughput-rtx_pro_6000_blackwell_server_edition}
\vspace{5pt}
\begin{tabular}{@{}llrrrrrrr@{\hspace{10pt}}rrrrrrr@{}}
\toprule
 & & \multicolumn{7}{c}{Forward (TFLOP/s $\uparrow$)} & \multicolumn{7}{c}{Backward (TFLOP/s $\uparrow$)} \\
\cmidrule(lr){3-9} \cmidrule(l){10-16}
 & System & 128 & 256 & 512 & 1K & 2K & 4K & 8K & 128 & 256 & 512 & 1K & 2K & 4K & 8K \\
\midrule
Shaw & PyTorch eager & 23 & 23 & 18 & 15 & 12 & 8.6 & 5.8 & 32 & 36 & 27 & 22 & 16 & 11 & 6.6 \\
 & PyTorch compile & 51 & 49 & 37 & 31 & 27 & 23 & 17 & 49 & 57 & 17 & 5.1 & 2.0 & 0.88 & 0.41 \\
 & PyTorch SDPA & 53 & 57 & 42 & 30 & 21 & 13 & 7.5 & \textbf{51} & 61 & 45 & 33 & 22 & 13 & 7.4 \\
 & FlexAttention & 47 & 46 & 33 & 25 & 18 & 12 & 7.1 & 39 & 44 & 32 & 24 & 18 & 11 & 6.8 \\
 & FA4 & 44 & 38 & 26 & 20 & 15 & 10 & 6.5 & -- & -- & -- & -- & -- & -- & -- \\
 & FlashInfer & 46 & 44 & 29 & 21 & 16 & 11 & 6.7 & -- & -- & -- & -- & -- & -- & -- \\
\cmidrule{2-16}
 & FlexAttention + decomp. & 60 & \textbf{88} & 89 & 117 & 141 & 160 & 173 & 30 & 33 & 37 & 40 & 44 & 44 & 47 \\
 & FA4 + decomp. & 54 & 61 & 68 & 84 & 96 & 105 & 109 & -- & -- & -- & -- & -- & -- & -- \\
 & FlashInfer + decomp. & 51 & 60 & 37 & 43 & 51 & 58 & 63 & -- & -- & -- & -- & -- & -- & -- \\
 & \textbf{\ourslabel} & \textbf{65} & 81 & \textbf{95} & \textbf{134} & \textbf{184} & \textbf{227} & \textbf{260} & 47 & \textbf{63} & \textbf{69} & \textbf{91} & \textbf{112} & \textbf{127} & \textbf{140} \\
\midrule
TXL & PyTorch eager & 9.8 & 11 & 12 & 12 & 13 & 13 & 13 & 15 & 15 & 17 & 18 & 18 & 19 & 19 \\
 & PyTorch compile & 33 & 35 & 37 & 46 & 49 & 52 & 53 & 30 & 36 & 28 & 34 & 38 & 41 & 41 \\
 & PyTorch SDPA & 13 & 14 & 16 & 17 & 18 & 18 & 17 & 18 & 19 & 21 & 21 & 22 & 23 & 23 \\
 & FlexAttention & 13 & 15 & 16 & 17 & 18 & 18 & 17 & 16 & 17 & 18 & 19 & 19 & 20 & 20 \\
 & FA4 & 12 & 14 & 15 & 15 & 16 & 16 & 16 & -- & -- & -- & -- & -- & -- & -- \\
 & FlashInfer & 12 & 14 & 15 & 16 & 17 & 17 & 16 & -- & -- & -- & -- & -- & -- & -- \\
 & NeMo FlexAttention & 42 & 46 & 56 & 63 & 68 & 71 & 74 & 27 & 31 & 26 & 31 & 34 & 35 & 34 \\
\cmidrule{2-16}
 & FlexAttention + decomp. & 60 & \textbf{88} & \textbf{111} & \textbf{132} & \textbf{147} & \textbf{154} & \textbf{159} & 28 & 31 & 38 & 41 & 43 & 43 & 41 \\
 & FA4 + decomp. & 54 & 61 & 71 & 87 & 93 & 101 & 102 & -- & -- & -- & -- & -- & -- & -- \\
 & FlashInfer + decomp. & 50 & 60 & 64 & 68 & 68 & 71 & 72 & -- & -- & -- & -- & -- & -- & -- \\
 & NeMo FlexAttention + decomp. & 60 & \textbf{88} & \textbf{111} & \textbf{132} & \textbf{147} & \textbf{154} & \textbf{159} & 28 & 31 & 38 & 41 & 43 & 43 & 41 \\
 & \textbf{\ourslabel} & \textbf{66} & 81 & 107 & 128 & 142 & 145 & 149 & \textbf{46} & \textbf{62} & \textbf{80} & \textbf{90} & \textbf{96} & \textbf{96} & \textbf{95} \\
\midrule
DeBERTa & PyTorch eager & 13 & 12 & 12 & 10 & 9.0 & 6.6 & 6.2 & 20 & 19 & 19 & 16 & 14 & 12\rlap{\textsuperscript{\textdagger}} & 11\rlap{\textsuperscript{\textdagger}} \\
 & PyTorch compile & 29 & 32 & 32 & 28 & 37 & 36 & 35 & 18 & 15 & 12 & 2.8 & 6.1 & 5.4\rlap{\textsuperscript{\textdagger}} & 5.2\rlap{\textsuperscript{\textdagger}} \\
 & FlexAttention & 17 & 17 & 17 & 15 & 13 & 9.1 & 8.5 & 23 & 21 & 21 & 17 & 15 & 12\rlap{\textsuperscript{\textdagger}} & 12\rlap{\textsuperscript{\textdagger}} \\
 & FA4 & 16 & 16 & 15 & 13 & 12 & 8.2 & 7.8 & -- & -- & -- & -- & -- & -- & -- \\
 & FlashInfer & 17 & 16 & 16 & 14 & 12 & 8.5 & 8.1 & -- & -- & -- & -- & -- & -- & -- \\
 & FlashDeBERTa & 16 & 24 & 31 & 36 & 41 & 47 & 50 & 13 & 14 & 12 & 4.4 & 2.2 & 1.6\rlap{\textsuperscript{\textdagger}} & 1.6\rlap{\textsuperscript{\textdagger}} \\
 & GLiNER4j & \textbf{58} & 57 & 49 & 39 & 32 & 29 & 31 & -- & -- & -- & -- & -- & -- & -- \\
\cmidrule{2-16}
 & FlexAttention + decomp. & 54 & \textbf{62} & \textbf{62} & 61 & 80 & 95 & 106 & 20 & 18 & 20 & 22 & 23 & 23 & 23 \\
 & FA4 + decomp. & 43 & 46 & 54 & 59 & 66 & 71 & 77 & -- & -- & -- & -- & -- & -- & -- \\
 & FlashInfer + decomp. & 33 & 33 & 34 & 32 & 33 & 35 & 34 & -- & -- & -- & -- & -- & -- & -- \\
 & \textbf{\ourslabel} & 55 & 61 & 60 & \textbf{62} & \textbf{88} & \textbf{124} & \textbf{159} & \textbf{44} & \textbf{47} & \textbf{45} & \textbf{44} & \textbf{61} & \textbf{81} & \textbf{98} \\
\midrule
T5 & PyTorch eager & 15 & 16 & 18 & 19 & 18 & 8.7 & 8.4 & 20 & 22 & 22 & 22 & 20 & 15 & 10.0 \\
 & PyTorch compile & 37 & 40 & 35 & 39 & 43 & 45 & 46 & 23 & 25 & 17 & 8.1 & 3.9 & 1.9 & 0.93 \\
 & PyTorch SDPA & 4.9 & 5.7 & 6.7 & 7.3 & 6.9 & 5.3 & 5.1 & 8.7 & 11 & 13 & 13 & 13 & 10 & 7.6 \\
 & FlexAttention & 57 & 105 & 143 & 146 & 138 & 37 & 20 & \textbf{50} & \textbf{72} & 78 & 65 & 43 & 24 & 12 \\
 & FA4 & 44 & 58 & 68 & 71 & 60 & 25 & 16 & -- & -- & -- & -- & -- & -- & -- \\
 & FlashInfer & 50 & 79 & 96 & 98 & 80 & 31 & 18 & -- & -- & -- & -- & -- & -- & -- \\
\cmidrule{2-16}
 & FlexAttention + decomp. & \textbf{67} & \textbf{111} & 144 & 171 & 183 & 192 & 197 & 8.3 & 2.1 & 1.4 & 1.1 & 0.92 & 0.97 & 0.86 \\
 & FA4 + decomp. & 57 & 79 & 115 & 142 & 168 & 179 & 186 & -- & -- & -- & -- & -- & -- & -- \\
 & FlashInfer + decomp. & 63 & 109 & 145 & 166 & 190 & 199 & 203 & -- & -- & -- & -- & -- & -- & -- \\
 & \textbf{\ourslabel} & 64 & \textbf{111} & \textbf{169} & \textbf{224} & \textbf{261} & \textbf{283} & \textbf{296} & 36 & 60 & \textbf{92} & \textbf{117} & \textbf{135} & \textbf{148} & \textbf{154} \\
\bottomrule
\end{tabular}
\end{table*}

\begin{table*}[t]
\centering
\footnotesize
\setlength{\tabcolsep}{3pt}
\caption{Throughput of every system on the RTX A6000 (FP16) at every length. Engines running our decomposition use their faster variant per mechanism and pass. \textbf{Bold}: the fastest system. \textdagger: fails the FP32 check. --: not supported.}
\label{tab:throughput-rtx_a6000}
\vspace{5pt}
\begin{tabular}{@{}llrrrrrrr@{\hspace{10pt}}rrrrrrr@{}}
\toprule
 & & \multicolumn{7}{c}{Forward (TFLOP/s $\uparrow$)} & \multicolumn{7}{c}{Backward (TFLOP/s $\uparrow$)} \\
\cmidrule(lr){3-9} \cmidrule(l){10-16}
 & System & 128 & 256 & 512 & 1K & 2K & 4K & 8K & 128 & 256 & 512 & 1K & 2K & 4K & 8K \\
\midrule
Shaw & PyTorch eager & 7.5 & 9.0 & 7.5 & 6.2 & 5.3 & 4.1 & 3.0 & 11 & 13 & 11 & 8.0 & 5.4 & 3.3 & 1.8 \\
 & PyTorch compile & 12 & 16 & 14 & 12 & 11 & 8.9 & 6.7 & 16 & 21 & 11 & 4.0 & 1.7 & 0.75 & 0.35 \\
 & PyTorch SDPA & 17 & 20 & 16 & 13 & 10 & 6.9 & 4.3 & 14 & 16 & 12 & 9.1 & 6.1 & 3.6 & 2.0 \\
 & FlexAttention & 14 & 17 & 14 & 11 & 8.8 & 6.2 & 4.0 & 13 & 16 & 12 & 8.6 & 5.5 & 3.3 & 1.8 \\
 & FlashInfer & 13 & 15 & 11 & 8.6 & 7.1 & 5.3 & 3.6 & -- & -- & -- & -- & -- & -- & -- \\
\cmidrule{2-16}
 & FlexAttention + decomp. & 25 & 35 & 38 & 45 & 51 & 56 & 58 & 11 & 12 & 16 & 19 & 20 & 21 & 21 \\
 & FlashInfer + decomp. & 20 & 26 & 17 & 21 & 25 & 29 & 31 & -- & -- & -- & -- & -- & -- & -- \\
 & Flashlight + decomp. & 20 & 27 & 35 & 40 & 43 & 45 & 46 & -- & -- & -- & -- & -- & -- & -- \\
 & \textbf{\ourslabel} & \textbf{26} & \textbf{36} & \textbf{39} & \textbf{54} & \textbf{66} & \textbf{76} & \textbf{83} & \textbf{17} & \textbf{23} & \textbf{25} & \textbf{32} & \textbf{39} & \textbf{45} & \textbf{49} \\
\midrule
TXL & PyTorch eager & 3.8 & 4.4 & 4.7 & 4.9 & 5.0 & 5.0 & 4.8 & 5.2 & 6.1 & 6.7 & 7.0 & 7.1 & 6.8 & 6.6 \\
 & PyTorch compile & 9.4 & 13 & 16 & 18 & 19 & 19 & 19 & 8.8 & 8.9 & 11 & 13 & 14 & 18 & 16 \\
 & PyTorch SDPA & 5.3 & 5.9 & 6.4 & 6.7 & 6.7 & 6.7 & 6.3 & 5.9 & 6.6 & 7.2 & 7.5 & 7.8 & 7.7 & 7.5 \\
 & FlexAttention & 5.4 & 6.1 & 6.6 & 6.9 & 6.9 & 6.9 & 6.6 & 5.6 & 6.5 & 7.1 & 7.3 & 7.2 & 6.9 & 6.7 \\
 & FlashInfer & 5.1 & 5.8 & 6.0 & 6.2 & 6.2 & 6.2 & 5.9 & -- & -- & -- & -- & -- & -- & -- \\
 & NeMo FlexAttention & 13 & 18 & 22 & 24 & 25 & 26 & 25 & 8.3 & 8.1 & 9.8 & 11 & 11 & 14 & 13 \\
\cmidrule{2-16}
 & FlexAttention + decomp. & 25 & 35 & 42 & \textbf{50} & 53 & \textbf{55} & \textbf{52} & 10 & 11 & 14 & 15 & 16 & 16 & 16 \\
 & FlashInfer + decomp. & 20 & 26 & 30 & 32 & 33 & 33 & 34 & -- & -- & -- & -- & -- & -- & -- \\
 & NeMo FlexAttention + decomp. & 25 & 35 & 42 & \textbf{50} & \textbf{54} & \textbf{55} & \textbf{52} & 10 & 11 & 14 & 15 & 16 & 16 & 16 \\
 & Flashlight + decomp. & 12 & 18 & 23 & 27 & 28 & 29 & 27 & -- & -- & -- & -- & -- & -- & -- \\
 & \textbf{\ourslabel} & \textbf{26} & \textbf{36} & \textbf{43} & \textbf{50} & 53 & 54 & \textbf{52} & \textbf{17} & \textbf{23} & \textbf{27} & \textbf{30} & \textbf{32} & \textbf{34} & \textbf{34} \\
\midrule
DeBERTa & PyTorch eager & 5.5 & 5.6 & 5.5 & 3.9 & 3.3 & 2.7 & 2.3 & 6.6 & 7.8 & 8.1 & 6.6 & 6.1 & 5.6\rlap{\textsuperscript{\textdagger}} & 5.0\rlap{\textsuperscript{\textdagger}} \\
 & PyTorch compile & 9.5 & 13 & 14 & 13 & 12 & 13 & 12 & 7.8 & 8.2 & 6.3 & 2.0 & 3.1 & 1.4\rlap{\textsuperscript{\textdagger}} & 3.3\rlap{\textsuperscript{\textdagger}} \\
 & FlexAttention & 7.5 & 7.7 & 7.9 & 5.5 & 4.7 & 3.7 & 3.1 & 7.0 & 8.4 & 8.7 & 7.0 & 6.2 & 5.7\rlap{\textsuperscript{\textdagger}} & 5.2\rlap{\textsuperscript{\textdagger}} \\
 & FlashInfer & 7.2 & 7.3 & 7.2 & 5.0 & 4.2 & 3.4 & 2.8 & -- & -- & -- & -- & -- & -- & -- \\
 & FlashDeBERTa & 7.0 & 11 & 14 & 16 & 17 & 19 & 20 & 5.6 & 6.5 & 5.7 & 2.3 & 1.2 & 0.88\rlap{\textsuperscript{\textdagger}} & 0.85\rlap{\textsuperscript{\textdagger}} \\
 & GLiNER4j & \textbf{29} & \textbf{27} & \textbf{23} & 19 & 16 & 13 & 13 & -- & -- & -- & -- & -- & -- & -- \\
\cmidrule{2-16}
 & FlexAttention + decomp. & 22 & 25 & \textbf{23} & 23 & 29 & 33 & 36 & 8.5 & 7.6 & 8.6 & 11 & 11 & 11 & 10 \\
 & FlashInfer + decomp. & 15 & 15 & 16 & 16 & 17 & 18 & 19 & -- & -- & -- & -- & -- & -- & -- \\
 & Flashlight + decomp. & 18 & 20 & \textbf{23} & \textbf{25} & 26 & 27 & 27 & -- & -- & -- & -- & -- & -- & -- \\
 & \textbf{\ourslabel} & 23 & 26 & \textbf{23} & 23 & \textbf{32} & \textbf{42} & \textbf{52} & \textbf{15} & \textbf{16} & \textbf{15} & \textbf{15} & \textbf{21} & \textbf{27} & \textbf{33} \\
\midrule
T5 & PyTorch eager & 5.0 & 6.1 & 4.9 & 3.7 & 3.7 & 3.6 & 3.4 & 6.9 & 8.5 & 9.5 & 9.6 & 9.1 & 7.7 & 5.3 \\
 & PyTorch compile & 8.4 & 12 & 15 & 16 & 17 & 18 & 19 & 8.8 & 12 & 9.1 & 5.6 & 3.0 & 1.6 & 0.79 \\
 & PyTorch SDPA & 1.7 & 2.2 & 2.0 & 2.0 & 2.1 & 2.1 & 2.0 & 3.1 & 4.2 & 4.8 & 5.0 & 4.9 & 4.2 & 3.3 \\
 & FlexAttention & 29 & 53 & 55 & 39 & 26 & 15 & 8.0 & \textbf{22} & \textbf{30} & \textbf{32} & 27 & 17 & 11 & 6.8 \\
 & FlashInfer & 23 & 35 & 38 & 29 & 21 & 13 & 7.1 & -- & -- & -- & -- & -- & -- & -- \\
\cmidrule{2-16}
 & FlexAttention + decomp. & 32 & 52 & 57 & 56 & 59 & 60 & 62 & 5.8 & 1.5 & 0.94 & 0.84 & 0.72 & 0.68 & 0.63 \\
 & FlashInfer + decomp. & 28 & 38 & 41 & 43 & 44 & 47 & 45 & -- & -- & -- & -- & -- & -- & -- \\
 & Flashlight + decomp. & 36 & 51 & 56 & 57 & 57 & 57 & 56 & -- & -- & -- & -- & -- & -- & -- \\
 & \textbf{\ourslabel} & \textbf{37} & \textbf{62} & \textbf{74} & \textbf{83} & \textbf{89} & \textbf{93} & \textbf{96} & 14 & 24 & 31 & \textbf{38} & \textbf{46} & \textbf{48} & \textbf{50} \\
\bottomrule
\end{tabular}
\end{table*}

\section{Code and Data Availability}
\label{app:reproducibility}
We will open-source the code of \ours, our benchmark and end-to-end scripts, and all result files. The result files record every measurement together with its software versions, tile sizes, and GPU telemetry, and our scripts generate every result in this paper from them. The code includes tests that compare every output and gradient with an FP32 reference under padding, packing, causal masks, dropout, and each input precision. All models and datasets we use are public (\cref{app:experimental-details}).

\end{document}